\documentclass[10pt,twocolumn,letterpaper]{article}
\usepackage[svgnames,table]{xcolor}
\usepackage{iccv}
\usepackage{times}
\usepackage{epsfig}
\usepackage{graphicx}
\usepackage{gensymb} 
\usepackage{color}
\usepackage[pagebackref=true,breaklinks=true,letterpaper=true,colorlinks,bookmarks=false]{hyperref}
\usepackage{amsmath,amsfonts,amssymb,mathrsfs}
\usepackage{makecell}

\usepackage{graphicx}
\usepackage{amsmath,amssymb}
\usepackage{epstopdf}
\usepackage{tabularx}
\usepackage{eqnarray}
\usepackage{array,booktabs}
\usepackage{enumitem}
\usepackage[space]{grffile} 
\usepackage{soul} 
\usepackage{float}
\usepackage{tabularx,ragged2e} 
\usepackage{bbm}
\usepackage{pifont} 
\usepackage{cancel}
\usepackage{wrapfig}
\usepackage{color, colortbl}
\usepackage{tikz} 

\newcommand{\circled}[1]{\textcircled{#1}}

\makeatletter
\renewcommand{\paragraph}{%
  \@startsection{paragraph}{4}%
  {\z@}{0.3ex \@plus 1ex \@minus .2ex}{-1em}
  {\normalfont\normalsize\bfseries}%
}
\makeatother

\definecolor{myGray}{rgb}{0.6,0.6,0.6}
\definecolor{myBlue}{rgb}{0.1,0.1,0.8}

\newcommand{\RR}{\mathbb{R}}

\newcommand{\nbyn}[1]{$#1\hspace{-0.30em}\times\hspace{-0.30em}#1$}
\newcommand{\nbynloose}[1]{$#1\hspace{-0.15em}\times\hspace{-0.15em}#1$}

\newcommand{\nospace}{\hspace{-0.35em}}

\def\onedot{\ifx\@let@token.\else.\null\fi\xspace}

\def\eg{\emph{e.g}\onedot} 
\def\ie{\emph{i.e}\onedot} 
 
\def\etc{\emph{etc}\onedot} 
\def\etal{\emph{et al}\onedot}

\definecolor{pcGray}{rgb}{0.5,0.5,0.5}

\definecolor{pccGray}{rgb}{0.3,0.3,0.3}

\newcommand{\fig}{Fig.~}
\newcommand{\eq}{Eq.\,}
\newcommand{\sect}{Section~}

\newcommand\closedots{\makebox[1em][c]{.\hfil.\hfil.}}

\newcommand\closer{\hspace{-.2em}}

\usepackage{listings}
\usepackage{color}

\ifx \@etal \@empty \newcommand{\etal}{et al.} \fi
\ifx \@eg \@empty \newcommand{\eg}{e.g.,~} \fi
\ifx \@ie \@empty \newcommand{\ie}{i.e.,~} \fi
\ifx \@etc \@empty \newcommand{\etc}{etc.} \fi

\newcommand{\ba}{\boldsymbol{a}}
\newcommand{\bb}{\boldsymbol{b}}

\newcommand{\bh}{{\boldsymbol{h}}}

\newcommand{\bq}{{\boldsymbol{q}}}

\newcommand{\bv}{{\boldsymbol{v}}}
\newcommand{\bw}{{\boldsymbol{w}}}
\newcommand{\bx}{{\boldsymbol{x}}}
\newcommand{\by}{{\boldsymbol{y}}}

\def\thmcolon{\hspace{-.85em} {\bf :}}

\newtheorem{THEOREM}{Theorem}[section]
\newtheorem{LEMMA}[THEOREM]{Lemma}
\newtheorem{PROPOSITION}[THEOREM]{Proposition}
\newtheorem{COROLLARY}[THEOREM]{Corollary}
\newtheorem{DEFINITION}[THEOREM]{Definition}
\newtheorem{OBSERVATION}[THEOREM]{Observation}

\definecolor{bgCode}{rgb}{0.94, 0.94, 1.0}
\graphicspath{{figures/},{plots/}}

\iccvfinalcopy

\ificcvfinal\fi

\makeatletter
\let\@fnsymbol\@arabic
\makeatother

\begin{document}


\title{Tips and Tricks for Visual Question Answering:\\Learnings from the 2017 Challenge}

\author{%
  Damien Teney$^*$~~~~~~Peter Anderson$^\dag$\thanks{Work performed while interning at Microsoft.}~~~~~~Xiaodong He$^\ddag$~~~~~~Anton van den Hengel$^*$\vspace{5pt}\\
  $^*$Australian Centre for Visual Technologies, The University of Adelaide, Australia\\
  $^\dag$Australian National University, Canberra, Australia\\
  $^\ddag$Deep Learning Technology Center, Microsoft Research, Redmond, WA, USA\vspace{5pt}\\
  {\tt\small damien.teney@adelaide.edu.au, peter.anderson@anu.edu.au,}\vspace{-2pt}\\
  {\tt\small xiaohe@microsoft.com, anton.vandenhengel@adelaide.edu.au}
}

\maketitle

\begin{abstract}
This paper presents a state-of-the-art model for visual question answering (VQA), which won the first place in the 2017 \emph{VQA Challenge}. VQA is a task of significant importance for research in artificial intelligence, given its multimodal nature, clear evaluation protocol, and potential real-world applications. The performance of deep neural networks for VQA is very dependent on choices of architectures and hyperparameters. To help further research in the area, we describe in detail our high-performing, though relatively simple model. Through a massive exploration of architectures and hyperparameters representing more than 3,000 GPU-hours, we identified tips and tricks that lead to its success, namely: sigmoid outputs, soft training targets, image features from bottom-up attention, gated tanh activations, output embeddings initialized using GloVe and Google Images, large mini-batches, and smart shuffling of training data. We provide a detailed analysis of their impact on performance to assist others in making an appropriate selection.
\end{abstract}

\section{Introduction}

\begin{figure}[t]
  \centering
  \renewcommand{\tabcolsep}{0.0em}
  \renewcommand{\arraystretch}{0.8}
  \begin{tabular}{cccc}
    \includegraphics[height=0.16\textwidth]{candles.jpg}&
    \includegraphics[height=0.16\textwidth]{hydrant.jpg}\\
    \scriptsize{\textbf{\texttt{What is on the coffee table ?~~}}}&
    \scriptsize{\textbf{\texttt{~What color is the hydrant ?}}}\\
    \scriptsize{\textbf{\textit{\texttt{candles}}}}&
    \vspace{1em}\scriptsize{\textbf{\textit{\texttt{black and yellow}}}}\\
    \includegraphics[height=0.16\textwidth]{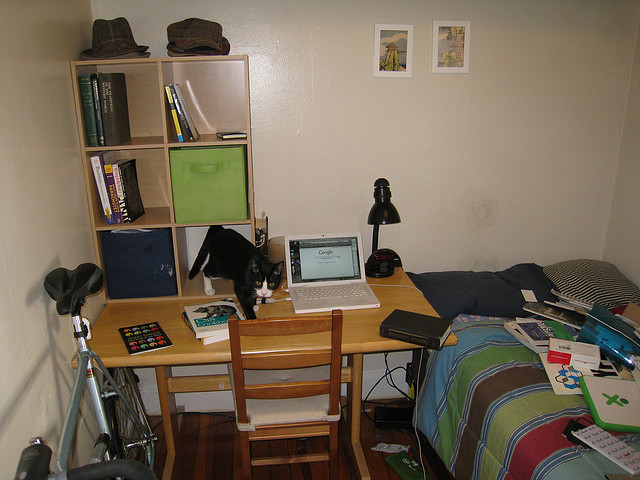}&
    \includegraphics[height=0.16\textwidth]{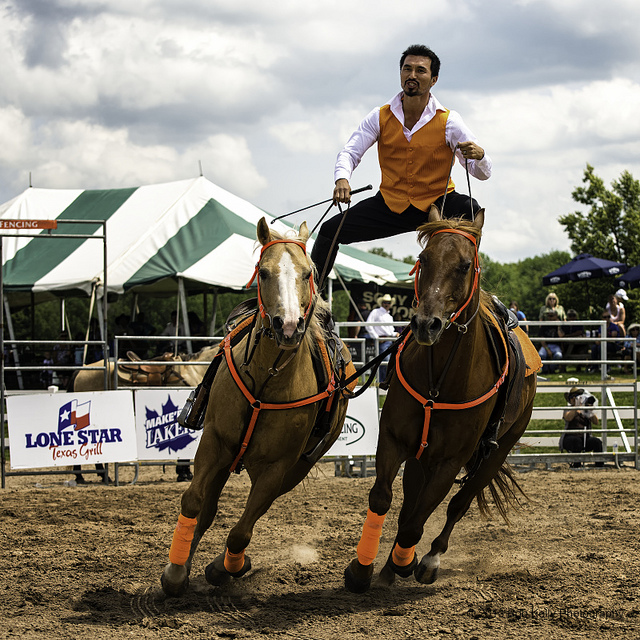}\\
    \scriptsize{\textbf{\texttt{What is on the bed ?}}}&
    \scriptsize{\textbf{\texttt{What is the long stick for ?}}}\\
    \scriptsize{\textbf{\texttt{\textit{books}}}}&
    \scriptsize{\textbf{\texttt{\textit{whipping}}}}
  \end{tabular}
  \label{fig:vqa}
  \caption{The task of visual question answering (VQA) relates visual concepts with elements of language and, occasionally, common-sense or general knowledge. Examples of training questions and their correct answer from the \emph{VQA v2} dataset~\cite{goyal2016balanced}.}
\end{figure}

The task of Visual Question Answering (VQA) involves an image and a related text question, to which the machine must determine the correct answer~(see \fig~\ref{fig:vqa}). The task lies at the intersection of the fields of computer vision, natural language processing, and artificial intelligence. This paper presents a relatively simple model for VQA that achieves state-of-the-art results. It is based on a deep neural network that implements the well-known joint embedding approach. The details of its architecture and hyperparameters were carefully selected for optimal performance on the VQA v2 benchmark~\cite{goyal2016balanced}. Admittedly, a large part of such a search is necessarily guided by empirical exploration and validation. Given the limited understanding of neural networks trained on a task as complex as VQA, small variations of hyperparameters and of network architectures may have significant and sometimes unpredictable effects on final performance~\cite{kazemi2017baseline}. The aim of this paper is to share the details of a successful model for VQA. The findings reported herein may serve as a basis for future development of VQA systems and multimodal reasoning algorithms in general. 

The proposed model is based on the principle of a joint embedding of the input question and image, followed by a multi-label classifier over a set of candidate answers. While this general approach forms the basis of many modern VQA methods~\cite{wu2017survey}, the details of the model are critical to achieving a high quality result. We also complement our model with a few key technical innovations that greatly enhance its performance. We have conducted an extensive empirical study to explore the space of architectures and hyperparameters to determine the importance of the various components.

\subsection{Summary of findings}
Our key findings are summarized with the following characteristics of the proposed model, which enable its high performance (see also Table~\ref{tableCumul}).
\setlist[itemize]{leftmargin=*}
\begin{itemize}[label={--}]
  \setlength\itemsep{4pt}
  \leftmargin=-10pt
  \itemindent=0pt
  \listparindent=0pt
  \topsep=4pt

  \item Using a \textbf{sigmoid output} that allows multiple correct answers per question, instead of a common single-label softmax.

  \item Using \textbf{soft scores as ground truth targets} that cast the task as a regression of scores for candidate answers, instead of a traditional classification.

  \item Using \textbf{gated tanh activations} in all non-linear layers.

  \item Using \textbf{image features from bottom-up attention}~\cite{anderson2017features} that provide region-specific features, instead of traditional grid-like feature maps from a CNN.

  \item Using \textbf{pretrained representations of candidate answers} to initialize the weights of the output layer.

  \item Using \textbf{large mini-batches} and \textbf{smart shuffling} of training data during stochastic gradient descent.

\end{itemize}


\begin{figure*}[t]
  \centering
  \includegraphics[width=0.99\textwidth]{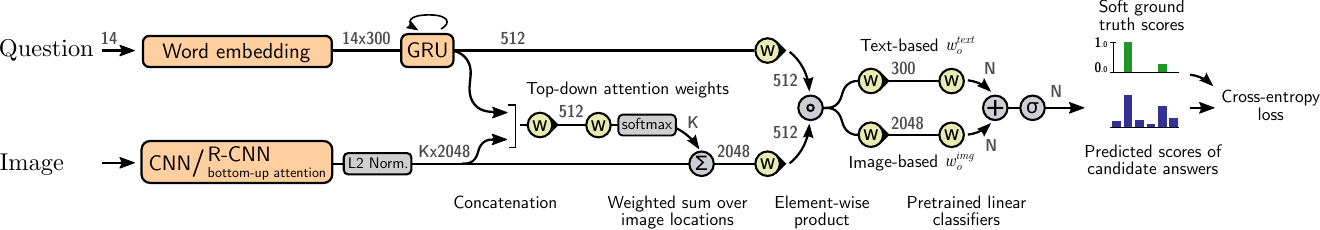}
  \label{fig:network}
  \caption{Overview of the proposed model. A deep neural network implements a joint embedding of the input question and image, followed by a multi-label classifier over a fixed set of candidate answers. Gray numbers indicate the dimensions of the vector representations between layers. Yellow elements use learned parameters. The elements \circled{$\hspace{.2em}\mathsf{w}\hspace{.1em}$} represent linear layers, and \circled{$\hspace{.2em}\mathsf{w}\hspace{.1em}$}\hspace{-.23em}$\rhd$ non-linear layers (gated tanh).}
\end{figure*}

\section{Background}

The task of VQA has gathered increasing interest in the past few years since the seminal paper of Antol \etal~\cite{antol2015vqa}. Even though the task straddles the fields of computer vision and natural language processing, it has primarily been a focus of the former.  This is partly because VQA constitutes a practical setting to evaluate deep visual understanding, itself considered the over-arching goal of computer vision. The task of VQA is extremely challenging, since it requires the comprehension of a text question, the parsing of visual elements of an image, and reasoning over those modalities, sometimes on the basis of external or common-sense knowledge~(see \fig\ref{fig:vqa}). The increasing interest in VQA parallels a similar trend for other tasks involving vision and language, such as image captioning~\cite{fang2014captions,vinyals2014show} and visual dialog~\cite{das2017dialog}.

\paragraph{Datasets} A number of large-scale datasets for VQA have been created~(\eg~\cite{antol2015vqa,goyal2016balanced,krishnavisualgenome,zhu2015visual7w}; see~\cite{wu2017survey} for a survey). Each dataset contains various images, typically from Flickr and/or from the COCO dataset~\cite{lin2014microsoft}, together with human-proposed questions and ground truth answers. The \emph{VQA-real} dataset of Antol \etal.~\cite{antol2015vqa} has served as the \emph{de facto} benchmark since its introduction in 2015. As the performance of methods improved, however, it became apparent that language-based priors and rote-learning of example questions/answers were overly effective ways to obtain good performance~\cite{goyal2016balanced,jabri2016revisiting,zhang2015balanced}. That fact hinders the effective evaluation and comparison of competing methods. The observation led to the introduction of a new version of the dataset, referred to as \emph{VQA v2}~\cite{goyal2016balanced}. It associates two images to every question. Crucially, the two images are chosen so as to each lead to different answers. This obviously discourages blind guesses, \ie inferring the answer from the question alone. This new setting and dataset were the basis of the 2017 VQA challenge~\cite{vqaleaderboard} and of the experiments presented in this paper.

Another dataset used in this paper is the \emph{Visual Genome}~\cite{krishnavisualgenome}. This multipurpose dataset contains annotations of images in the form of scene graphs. Those constitute fine-grained descriptions of the image contents. They provide a set of visual elements appearing in the scene (\eg objects, persons), together with their attributes (\eg color, appearance) and the relations between them. We do not use these annotations directly, but they serve in~\cite{anderson2017features} to train a Faster R-CNN model~\cite{ren2015faster}, which we use here to obtain object-centric image features. We directly use other annotations of the \emph{Visual Genome} dataset which are simply questions relating to the images. In comparison to the questions of \emph{VQA v2}, these have more diverse formulations and a more varied set of answers. Those answers are also often longer, \ie short phrases, whereas most answers in \emph{VQA v2} are usually $1$ to $3$-words long. As described in Sect.\ref{sect:training}, we only use the subset of questions whose answers overlap those in \emph{VQA v2}.

\paragraph{Methods}
The prevailing approach to VQA is based on three components. \emph{(1)}~Posing question answering as a classification problem, solved with \emph{(2)}~a deep neural network that implements a joint embedding model, \emph{(3)}~trained end-to-end with supervision of example questions/answers. First, question-answering is posed as a classification over a set of candidate answers. Questions in the current VQA datasets are mostly visual in nature, and the correct answers therefore only span a small set of words and phrases. Practically, correct answers are concentrated in a small set of words and phrases (typically a few hundreds to a few thousands). Second, most VQA models are based on a deep neural network that implements a joint embedding of the image and of the question. The two inputs are mapped into fixed-size vector representations with convolutional and recurrent neural networks, respectively. Further non-linear mappings of those representations are usually interpreted as projections into a joint ``semantic'' space. They can then be combined by means of concatenation of element-wise multiplication, before feeding the classifier mentioned above. Third, owing to the success of deep learning on supervised learning problems, this whole neural network is trained end-to-end from questions, images, and their ground truth answers. Note that this constitutes a relatively sparse training signal considering the huge dimensionality of the input space of possible images and questions. This in turn requires massive amounts of training examples. This has driven the efforts in collecting large-scale datasets such as \emph{VQA v2}. It contains in the order of 650,000 questions with ground-truth answers, relating to about 120,000 different images.

The majority of methods proposed for VQA in the past few years have built on the basic joint embedding approach with more complex  mechanisms for attention (see~\cite{wu2017survey} for a recent survey). Interestingly, recent studies have shown that very simple models can also achieve strong performance~\cite{jabri2016revisiting,kazemi2017baseline} given careful implementation and/or selection of hyperparameters. Our work follows a similar line. Our extensive series of experiments show that a few keys choices in the implementation (\eg gated activations, regression output, smart shuffling, \etc) dramatically improve the performance of a relatively simple model. We also show how the performance of a naive implementation gradually improves with each of those choices. We therefore hope that the proposed model can serve as a strong basis for other future incremental developments.

\section{Proposed model}

This section presents the proposed model, based on a deep neural network. For the sake of transparency, we pragmatically describe the model with the specific choices and values of hyperparameters that lead to its best performance. Section \ref{sec:details} will examine variations of architecture and hyperparameters and their influence on performance.

The complete model is summarized in \fig\ref{fig:network}. As a one-sentence summary, it implements the well-known joint RNN/CNN embedding of the question/image, with question-guided attention over the image~\cite{wu2017survey,yang2015stacked,jabri2016revisiting,kazemi2017baseline}.

\subsection{Question embedding} The input for each instance --~whether during training or test time~-- is a text question and an image. The question is tokenized, \ie first split into words using spaces and punctuation. Any number or number-based word (\eg \textit{10,000} or \textit{2:15pm}) is also considered as a word. Questions are trimmed to a maximum of 14 words for computational efficiency. The extra words are then simply discarded. but only about 0.25\% of questions in the dataset are longer than 14 words. Each word is turned into a vector representation with a look-up table, whose entries are 300-dimensional vectors learned along other parameters during training. Those vectors are however initialized with pretrained \textit{GloVe} word embeddings~\cite{pennington2014glove} (Global Vectors for Word Representation). We use the publicly available version of GloVe pretrained on the Wikipedia/Gigaword corpus\footnote{\url{http://nlp.stanford.edu/projects/glove/}}. The words not present in the pretrained word embdding are initialized with vectors of zeros (subsequenty optimized during training). The questions shorter than 14 words are end-padded with vectors of zeros (frozen during training). The resulting sequence of word embeddings is of size $14\times300$ and it is passed through a Recurrent Gated Unit (GRU~\cite{cho2014learning}). The recurrent unit has an internal state of dimension 512, and we use its final state, \ie after processing the 14 word embeddings, as our question embedding $\bq$. Note that we do not use sentinel \texttt{start} or \texttt{end} tokens, nor do we trim sequences, or process the strict number of tokens in the given sentence (also known as per-example dynamic unrolling in~\cite{kazemi2017baseline}, or TrimZero in~\cite{Kim2016TrimZero}). We rather found it more effective to always run the recurrent units for the same number of iterations, including entries containing zero-padding.

\subsection{Image features} The input image is passed through a Convolutional Neural Network (CNN) to obtain a vector representation of size $K \times 2048$, where $K$ is a number of image locations. Each location is thus represented by a 2048-dimensional vector that encodes the appearance of the image in that region. Our evaluation in Section~\ref{sec:details} compares two main options with different trade-offs in commodity and performance: a standard pretrained CNN, or a better-performing option. The first, lower-performance option is a 200-layer ResNet (Residual Network~\cite{he2015resnet}) pretrained on ImageNet and publicly available~\cite{he2016identityresnet}. This gives feature maps of size \nbyn{14} that we resize by average pooling (\ie bilinear interpolation) to \nbyn{7} (\ie $K$=$49$). The second, higher-performance option is to use the method proposed in~\cite{anderson2017features} which provides image features using \emph{bottom-up attention}. The method is based on a ResNet CNN within a Faster~R-CNN framework~\cite{ren2015faster}. It is trained to focus on specific elements in the given image, using annotations from the \emph{Visual Genome} dataset~\cite{krishnavisualgenome}. The resulting features can be interpreted as ResNet features centered on the top-$K$ objects in the image. Our experiments evaluate both a fixed $K$=36, and an adaptive $K$ that uses a fixed threshold for the detected elements in the image, allowing the number of regions $K$ to vary with the complexity of each image, up to a maximum of $100$. The images used by the \emph{VQA v2} dataset yield in that case an average of about $K$=$60$ per image.

In all cases, the CNN is pretrained and held fixed during the training of the VQA model. The features can therefore be extracted from the input images as a preprocessing step for efficiency.

\subsection{Image attention} Our model implements a classical question-guided attention mechanism common to most modern VQA models (see \eg~\cite{zhu2015visual7w,xu2015ask,Chen2015ABC,Jiang2015compositional,andreas2015deep,yang2015stacked}). We refer to this stage as the \emph{top-down} attention, as opposed to the model of Anderson \etal. \cite{anderson2017features} that provides image features from bottom-up attention.

For each location $i=1 \closedots K$ in the image, the feature vector $\bv_i$ is concatenated with the question embedding $\bq$ (see \fig\ref{fig:network}). They are both passed through a non-linear layer $f_a$ (see Section~\ref{sec:nonLinearLayers}) and a linear layer to obtain a scalar attention weight $\alpha_{i,t}$ associated with that location. Formally,
\begin{align}
   a_{i}               & ~~=~~ \bw_a  f_a( [\bv_i,\bq]) \label{eqn:att-vqa} \\
   \boldsymbol{\alpha} & ~~=~~ \textrm{softmax}\,(\ba) \label{eqn:norm} \\
   \hat{\bv}           & ~~=~~ \Sigma_{i=1}^{K}{\alpha_{i}\bv_i} \label{eqn:attended}
\end{align}
where $\bw_a$ is a learned parameter vector.
The attention weights are normalized over all locations with a softmax function (Eq. \ref{eqn:norm}). The image features from all locations are then weighted by the normalized values and summed (\eq\ref{eqn:attended}) to obtain a single $2048$-sized vector $\hat{\bv}$ representing the attended image.

Note that this attention mechanism is a simple one-glimpse, one-way attention, as opposed to more complex schemes of recent models (\eg stacked, multi-headed, or bidirectional attention~\cite{yang2015stacked,jabri2016revisiting,kazemi2017baseline,lu2016hierarchical}).

\subsection{Multimodal fusion} The representations of the question ($\bq$) and of the image ($\hat{\bv}$) are passed through non-linear layers and then combined with a simple Hadamard product (\ie element-wise multiplication):
\begin{align}
  \bh ~~=~~ f_q(\bq) ~\circ~ f_v(\hat{\bv})
\end{align}
The resulting vector $\bh$ is referred to as the joint embedding of the question and of the image, and is then fed to the output classifier.

\subsection{Output classifier} A set of candidate answers, that we refer to as the output vocabulary, is predetermined from all the correct answers in the training set that appear more than 8 times. This amounts to $N$=$3129$ candidate answers. We treat VQA as a multi-label classification task. Indeed, each training question in the \emph{VQA v2} dataset is associated with one or several answers, each labeled with soft accuracies in $[0,1]$. Multiple answers and accuracies in $(0,1)$ arise in case of disagreement between human annotators, particularly with ambiguous questions and multiple or synonymous correct answers~\cite{goyal2016balanced}. Moreover, in our case, some training questions (about $7\%$) have no correct answer within the selected output vocabulary. Those questions are not discarded however. We find them to provide a useful training signal, by driving towards zero the scores predicted for all candidates of the output vocabulary (see Section~\ref{sec:ablativeTrData}).

Our multi-label classifier passes the joint embedding $\bh$ through a non-linear layer $f_o$ then through a linear mapping $\bw_o$ to predict a score $\hat{s}$ for each of the $N$ candidates:
\begin{align}
  \label{eq:classifier1}
  \hat{s} ~~=~~ \sigma\big( \, \bw_o \;~f_o(\bh) \, \big)
\end{align}
where $\sigma$ is a sigmoid (logistic) activation function, and $\bw_o \in \RR^{N \times 512}$ is a learned weight matrix initialized as described below.

The sigmoid normalizes the final scores to $(0,1)$, which are followed by a loss similar to a binary cross-entropy, although we use \emph{soft} target scores. This final stage can be seen as a logistic regression that predicts the correctness of each candidate answer. Our objective function is
\begin{align}
  \label{eq:loss}
  L ~=~ -\closer\sum_i^M \sum_j^N {s}_{ij} \log(\hat{s}_{ij}) \;-\; (1\closer-\closer{s}_{ij}) \log(1\closer-\closer \hat{s}_{ij})
\end{align}
where the indices $i$ and $j$ run respectively over the $M$ training questions and $N$ candidate answers. The ground-truth scores ${s}$ are the aforementioned soft accuracies of ground truth answers. The above formulation proved to be much more effective than a softmax classifier as commonly used in other VQA models. The advantage of the above formulation is two-fold. First, the sigmoid outputs allow 
optimization for multiple correct answers per question~\cite{jabri2016revisiting,teney2016zsvqa} as is occasionally the case in the VQA v2 dataset. Second, the use of soft scores as targets provides a slightly richer training signal than binary targets, as they capture the occasional uncertainty in ground truth annotations.

\subsection{Pretraining the classifier} In the output stage described above (\eq\ref{eq:classifier1}), the score of a candidate answer $j$ is effectively determined by a dot product between our joint image-question representation $f_o(\bh)$ and the $j$th row of $\bw_o$. During training, an appropriate representation for each candidate answer is thus learned as a row of $\bw_o$. We propose to use prior information about the candidate answers from two sources to initialize the rows of $\bw_o$. On the one hand, we use linguistic information in the form of the GloVe word embeddings of the answer word (as described above for \emph{Question embedding}). When the answer cannot be matched exactly with a pretrained embedding, we use the closest match after spell checking, removing hyphenation, or keeping a single term from multi-word expressions. The corresponding vectors are placed in the matrix $\bw_o^{\textrm{text}}$.

We also exploit visual information gathered from images representing the candidate answers. We use \textit{Google Images} to automatically retrieve 10 photographs associated with each candidate answer. Those photographs are passed through a ResNet-101 CNN pretrained on ImageNet~\cite{he2015resnet}. The final mean-pooled features are extracted and averaged over the 10 photographs. The resulting 2048-sized vector of each candidate answer is placed in the corresponding row of a matrix $\bw_o^{\textrm{img}}$. Those visual representations are complementary to the linguistic ones obtained through word embeddings. They can also be obtained for any candidate answer, including multi-word expressions and rare words for which no word embeddings are available. On the downside, abstract words and expressions may not lead to informative visual representations~(see \sect\ref{sec:classifiers} and \fig\ref{fig:answers}).

We combine the prior representations $\bw_o^{\textrm{text}}$ and $\bw_o^{\textrm{img}}$ as follows, decomposing \eq\ref{eq:classifier1} into
\begin{align}
  \label{eq:classifier2}
  \hat{s} ~~=~~ \sigma\big( \, \bw_o^{\textrm{text}} \;~f_o^{\textrm{text}}(\bh) ~+~ \bw_o^{\textrm{img}} \;~f_o^{\textrm{img}}(\bh) \, \big)
\end{align}
where the non-linear transformations $f_o^{\textrm{text}}$ and $f_o^{\textrm{img}}$ bring $\bh$ to the appropriate dimensions, \ie 300 and 2048 respectively (see \fig\ref{fig:network}). The matrices $\bw_o^{\textrm{text}}$ and $\bw_o^{\textrm{img}}$ are fine-tuned with the remainder of the network using smaller relative learning rates, respectively 0.5 and 0.01 (determined through cross-validation).

\subsection{Non-linear layers} \label{sec:nonLinearLayers} The network described above uses multiple learned non-linear layers (see \fig\ref{fig:network}). A common implementation for such a layer would be an affine transformation followed by a Rectified Linear Unit (ReLU). In our implementation, each non-linear layer uses a gated hyperbolic tangent activation. That is, each of those layers implements a function $f_a:\bx \in \mathbb{R}^{m} \to \by \in \mathbb{R}^{n}$ with parameters $a$ defined as follows:
\begin{align}
  \label{eq:nonLinearLayer}
  & \tilde{\by} ~~=~~ \tanh\,(W \bx + \bb )\\
  & \boldsymbol{g} ~~=~~ \sigma( W' \bx + \bb' )\\
  & \by ~~=~~ \tilde{\by} ~\circ~ \boldsymbol{g}
\end{align}
where $\sigma$ is the sigmoid activation function, $W, W' \in \mathbb{R}^{n \times m}$ are learned weights, $\bb, \bb' \in \mathbb{R}^{n}$ are learned biases, and~$\circ$~is the Hadamard (element-wise) product. The vector $\boldsymbol{g}$ acts multiplicatively as a gate on the intermediate activation $\tilde{\by}$. That formulation is inspired by similar gating operations within recurrent units such as LSTMs and GRUs~\cite{cho2014learning}. This can also be seen as a special case of highway networks~\cite{srivastava2015highway} and has been mentioned in other work in natural language processing~\cite{dauphin2016languagecnn,teney2016graphvqa}.

\subsection{Training} \label{sect:training} We train the network using stochastic gradient descent. We use the AdaDelta algorithm~\cite{zeiler2012adadelta}, which does not require fixing learning rates and is very insensitive to the initialization of the parameters. The model is prone to overfitting, which we prevent by early stopping as follows. We first train leaving out the official validation set of the \emph{VQA v2} dataset for monitoring, and identify the epoch yielding the best performance (highest overall VQA score). The training is then repeated for the same number of epochs, now also using the validation set as training data (as in~\cite{goyal2016balanced}).

We use questions/answers from the \emph{Visual Genome}~\cite{krishnavisualgenome} as additional training data. We only use questions whose correct answers overlap the output vocabulary determined on the \emph{VQA v2} dataset. This amounts to only about $30\%$ or 485,000 questions from the \emph{Visual Genome}.

During training with stochastic gradient descent, we enforce the shuffling of training instance to keep \emph{balanced pairs} of \emph{VQA v2} in the same mini-batches. Those pairs correspond to identical questions with different images and answers. Our intuitive motivation is that such pairs likely lead to gradients pulling the network parameters in different directions. Keeping an example and its balanced counter-part in a same mini-batch is expected to make the learning more stable, and encourage the network to discern the subtle differences between the paired instances \cite{teney2016graphvqa}.

\subsection{Implementation} Our model is implemented entirely in Matlab using custom deep learning libraries, with the exception of some Java code for multithreaded loading of input data. Training one network usually converges in 12--18 epochs, which takes in the order of 12--18 hours with~$K=36$ on a single Nvidia K40 GPU, or about twice as long on a CPU.

\section{Ablative experiments}
\label{sec:details}

\rowcolors{4}{gray!25}{}

\begin{table*}[p]
\scriptsize
\renewcommand{\arraystretch}{1.28}
\centering
\begin{tabularx}{\textwidth}{Xrccccc}
\Xhline{1\arrayrulewidth}
\hiderowcolors
                                          &    &  \multicolumn{5}{c}{VQA v2 validation} \\ \cline{3-7}
                                          &    &  \multicolumn{4}{c}{VQA Score} & Accuracy over \\ \cline{3-6}
                                          &    &  All & Yes/no & Numbers & Other & balanced pairs \\
\hiderowcolors
\Xhline{1\arrayrulewidth}

\textbf{Reference model}  &    &  63.15 $\pm$0.08   &  80.07  &  42.87  &  55.81  &  34.66  \\
\hline

\multicolumn{6}{l}{\textbf{Training data} (reference: with Visual Genome data; shuffling keeps balanced pairs in the same minibatches)}\\
\showrowcolors
\hspace{1.5em} Without extra training data from Visual Genome &      &  62.48 $\pm$0.15   &  80.37  &  42.06  &  54.44  &  34.12  \\
\hspace{1.5em} Random shuffling of training data &        &  63.16 $\pm$0.06   &  80.17  &  42.92  &  55.73  &  34.54  \\
\hspace{1.5em} Discard training questions without answers of score=$1.0$ & &  63.15 $\pm$0.12   &  80.15  &  42.90  &  55.73  &  34.62  \\
\hline

\hiderowcolors\multicolumn{6}{l}{\textbf{Question embedding} (reference: 300-dimensional GloVe word embeddings, 1-layer forward GRU)}\\ \showrowcolors
\hspace{1.5em} Word embeddings learned from random initialization; 1-layer forward GRU &  &  62.28 $\pm$0.06   &  78.77  &  41.92  &  55.27  &  33.67  \\
\hspace{1.5em} 100-dimensional GloVe; forward GRU &            &  62.45 $\pm$0.05   &  78.84  &  42.12  &  55.51  &  33.70  \\
\hspace{1.5em} 200-dimensional GloVe; forward GRU &            &  62.96 $\pm$0.12   &  79.76  &  42.49  &  55.75  &  34.34  \\
\hspace{1.5em} 300-dimensional GloVe; bag-of-words (sum) &          &  62.14 $\pm$0.04   &  79.54  &  41.88  &  54.41  &  33.73  \\
\hspace{1.5em} 300-dimensional GloVe; bag-of-words (average) &         &  62.53 $\pm$0.09   &  80.17  &  42.15  &  54.67  &  34.22  \\
\hspace{1.5em} 300-dimensional GloVe; 1-layer backward GRU &         &  62.82 $\pm$0.02   &  79.57  &  42.39  &  55.64  &  33.98  \\
\hspace{1.5em} 300-dimensional GloVe; 2-layer forward GRU &          &  62.29 $\pm$0.10   &  78.66  &  42.42  &  55.23  &  33.56  \\
\hspace{1.5em} 300-dimensional GloVe randomly shuffled; 1-layer forward GRU &     &  62.16 $\pm$0.08   &  78.74  &  41.73  &  55.11  &  33.45  \\
\hline

\hiderowcolors\multicolumn{6}{l}{\textbf{Image features} (reference: image features from bottom-up attention, adaptive $K$)}\\ \showrowcolors
\hspace{1.5em} ResNet-200 global features ($K$=$1$, \ie without image attention)) &    &  56.16 $\pm$0.14   &  75.89  &  36.39  &  46.57  &  25.93  \\
\hspace{1.5em} ResNet-200 features \nbyn{14} ($K$=$196$)) &          &  57.93 $\pm$0.25   &  76.17  &  36.25  &  49.95  &  27.64  \\
\hspace{1.5em} ResNet-200 features downsampled to \nbyn{7} ($K$=$49$) &       &  59.35 $\pm$0.10   &  77.34  &  37.74  &  51.55  &  29.49  \\
\hspace{1.5em} Image features from bottom-up attention, $K$=$36$ &        &  62.82 $\pm$0.14   &  79.92  &  42.44  &  55.35  &  34.18  \\
\hline

\hiderowcolors\multicolumn{6}{l}{\textbf{Image attention} (reference: image features from bottom-up attention, 1 attention head, softmax normalization)}\\ \showrowcolors
\hspace{1.5em} ResNet-200 \nbyn{7} features, 1 head, sigmoid normalization &     &  58.96 $\pm$0.37   &  77.05  &  37.99  &  50.90  &  28.68  \\
\hspace{1.5em} ResNet-200 \nbyn{7} features, 1 head, softmax normalization &     &  59.35 $\pm$0.10   &  77.34  &  37.74  &  51.55  &  29.49  \\
\hspace{1.5em} ResNet-200 \nbyn{7} features, 2 heads, sigmoid normalization &     &  58.70 $\pm$0.30   &  76.42  &  37.95  &  50.87  &  28.40  \\
\hspace{1.5em} ResNet-200 \nbyn{7} features, 2 heads, softmax normalization &     &  59.20 $\pm$0.13   &  76.97  &  37.52  &  51.58  &  29.43  \\
\hspace{1.5em} Image features from bottom-up attention, 1 head, sigmoid normalization &   &  62.15 $\pm$1.41   &  78.83  &  43.44  &  54.57  &  32.98  \\
\hspace{1.5em} Image features from bottom-up attention, 2 heads, sigmoid normalization &  &  62.91 $\pm$0.26   &  79.74  &  43.99  &  55.27  &  34.30  \\
\hspace{1.5em} Image features from bottom-up attention, 2 heads, softmax normalization &  &  63.10 $\pm$0.05   &  79.80  &  43.17  &  55.82  &  34.52  \\
\hline

\hiderowcolors\multicolumn{6}{l}{\textbf{Output vocabulary} (reference: $>$8 occurrences as correct answers in the training data,~~$N=3,129$}\\ \showrowcolors
\hspace{1.5em} Keep answers with $>$6 training occurrences,~~$N=3,793$ &  &  63.26 $\pm$0.07   &  80.23  &  42.94  &  55.89  &  34.50  \\
\hspace{1.5em} Keep answers with $>$10 training occurrences,~~$N=2,748$ & $\ast$ &  63.30 $\pm$0.06   &  80.33  &  43.02  &  55.86  &  34.69  \\
\hspace{1.5em} Keep answers with $>$12 training occurrences,~~$N=2,418$ & &  63.16 $\pm$0.08   &  80.18  &  43.17  &  55.66  &  34.56  \\
\hspace{1.5em} Keep answers with $>$14 training occurrences,~~$N=2,160$ & &  63.27 $\pm$0.08   &  80.45  &  43.07  &  55.69  &  34.83  \\
\hspace{1.5em} Keep answers with $>$16 training occurrences,~~$N=1,961$ & &  63.17 $\pm$0.10   &  80.29  &  42.94  &  55.65  &  34.62  \\
\hspace{1.5em} Keep answers with $>$20 training occurrences,~~$N=1,656$ & &  63.14 $\pm$0.05   &  80.21  &  42.90  &  55.66  &  34.67  \\
\hspace{1.5em} Keep answers with $>$200 training occurrences,~~$N=278$ &  &  59.59 $\pm$0.04   &  79.90  &  42.11  &  48.93  &  32.64  \\
\hline

\hiderowcolors\multicolumn{6}{l}{\textbf{Output classifier} (reference: sigmoid output, $\bw_o^{\textrm{text}}$ and $\bw_o^{\textrm{img}}$ pretrained}\\ \showrowcolors

\hspace{1.5em} Softmax output, ~$\bw_o^{\textrm{text}}$ pretrained, ~~~~~~~~~~~~~~~~~$\bw_o^{\textrm{img}}$ pretrained &    &  60.47 $\pm$0.08   &  78.19  &  40.93  &  52.32  &  32.27  \\

\hspace{1.5em} Sigmoid output, ~$\bw_o^{\textrm{text}}$ randomly initialized, ~~$\bw_o^{\textrm{img}}$ randomly initialized &   &  62.28 $\pm$0.99   &  78.50  &  42.50  &  55.32  &  33.14  \\
\hspace{1.5em} Sigmoid output, ~$\bw_o^{\textrm{text}}$ randomly shuffled, ~~~~~$\bw_o^{\textrm{img}}$ randomly shuffled &     &  62.64 $\pm$0.15   &  79.82  &  42.26  &  55.13  &  34.17  \\

\hspace{1.5em} Sigmoid output, ~$\bw_o^{\textrm{text}}$ pretrained, ~~~~~~~~~~~~~~~~~~$\bw_o^{\textrm{img}}$ randomly initialized &  &  62.94 $\pm$0.14   &  79.99  &  42.72  &  55.47  &  34.26  \\
\hspace{1.5em} Sigmoid output, ~$\bw_o^{\textrm{text}}$ pretrained, ~~~~~~~~~~~~~~~~~~$\bw_o^{\textrm{img}}$ randomly shuffled &  &  63.04 $\pm$0.02   &  80.20  &  42.83  &  55.50  &  34.66  \\

\hspace{1.5em} Sigmoid output, ~$\bw_o^{\textrm{text}}$ randomly initialized, ~$\bw_o^{\textrm{img}}$ pretrained &      &  63.21 $\pm$0.01   &  80.06  &  42.97  &  55.90  &  34.65  \\
\hspace{1.5em} Sigmoid output, ~$\bw_o^{\textrm{text}}$ randomly shuffled, ~~~~$\bw_o^{\textrm{img}}$ pretrained &      &  62.97 $\pm$0.06   &  79.83  &  42.62  &  55.68  &  34.31  \\

\hline

\hiderowcolors\multicolumn{6}{l}{\textbf{General architecture} (reference: gated tanh non-linear activations, hidden states of dimension 512)}\\ \showrowcolors
\hspace{1.5em} Non-linear activations: ReLU &   &  61.63 $\pm$0.21   &  78.48  &  41.15  &  54.39  &  32.57  \\
\hspace{1.5em} Non-linear activations: tanh &   &  61.74 $\pm$0.44   &  79.20  &  40.92  &  54.13  &  33.10  \\
\hspace{1.5em} Non-linear activations: gated ReLU &  &  62.44 $\pm$0.07   &  79.33  &  42.12  &  55.13  &  33.52  \\
\hspace{1.5em} Hidden states of dimension 256 &   &  62.80 $\pm$0.05   &  79.62  &  42.46  &  55.53  &  34.12  \\
\hspace{1.5em} Hidden states of dimension 384 &   &  63.12 $\pm$0.11   &  80.06  &  42.89  &  55.74  &  34.54  \\
\hspace{1.5em} Hidden states of dimension 768 &   &  63.12 $\pm$0.11   &  80.06  &  42.89  &  55.74  &  34.54  \\
\hspace{1.5em} Hidden states of dimension 1024 &  &  63.02 $\pm$0.55   &  79.88  &  42.68  &  55.73  &  34.34  \\
\hspace{1.5em} Hidden states of dimension 1280 &$\ast$ &  63.37 $\pm$0.21   &  80.40  &  43.02  &  55.96  &  34.76  \\
\hline

\hiderowcolors\multicolumn{6}{l}{\textbf{Mini-batch size} (reference: 512 training questions)}\\ \showrowcolors
\hspace{1.5em} 128 Training questions &  &  62.38 $\pm$0.08   &  79.70  &  42.29  &  54.67  &  34.01  \\
\hspace{1.5em} 256 Training questions &$\ast$&  63.17 $\pm$0.09   &  80.22  &  42.94  &  55.72  &  34.71  \\
\hspace{1.5em} 384 Training questions &$\ast$&  63.20 $\pm$0.04   &  80.21  &  43.08  &  55.75  &  34.61  \\
\hspace{1.5em} 768 Training questions &  &  62.99 $\pm$0.12   &  79.84  &  42.61  &  55.73  &  34.40  \\

\Xhline{1\arrayrulewidth}
\end{tabularx}
\normalsize
\vspace{9pt}
\caption{\normalsize Ablations of a single network, evaluated on the VQA v2 validation set. We evaluate variations of our best ``reference'' model (first row), and show that it corresponds to a local optimum in the space of architectures and hyperparameters. Every row corresponds to one variation of that reference model. We train each variation with three different random seeds and report averages and standard deviations (\scriptsize$\pm$\normalsize).}
\label{tableAblative}
\end{table*}

\rowcolors{4}{gray!25}{}

\begin{table*}[t]
\scriptsize
\renewcommand{\arraystretch}{1.36}
\centering
\begin{tabularx}{\textwidth}{Xcccccc}
\Xhline{1\arrayrulewidth}
\hiderowcolors
                                          &    &  \multicolumn{5}{c}{VQA v2 validation} \\ \cline{3-7}
                                          &    &  \multicolumn{4}{c}{VQA Score} & Accuracy over \\ \cline{3-6}
                                          &    &  All & Yes/no & Numbers & Other & balanced pairs \\
\hiderowcolors
\Xhline{1\arrayrulewidth}

\textbf{Reference model}  &    &  63.15 $\pm$0.08   &  80.07  &  42.87  &  55.81  &  34.66  \\
\hline
\showrowcolors

(1)~~~Removing initialization of $\bw_o^{\textrm{text}}$ (randomly shuffled) &    &  63.01 $\pm$0.12   &  80.11  &  42.80  &  55.51  &  34.40  \\
(2)~~~Removing initialization of $\bw_o^{\textrm{img}}$ (randomly shuffled) &    &  62.67 $\pm$0.07   &  79.75  &  42.64  &  55.14  &  34.22  \\
(3)~~~Removing image features from bottom-up attention~\cite{anderson2017features}; \nbyn{7} ResNet-200 features instead &    &  58.90 $\pm$0.10   &  77.16  &  37.18  &  50.92  &  29.39  \\
(4)~~~Removing extra training data from the Visual Genome, only uses the VQA v2 training set &    &  57.84 $\pm$0.05   &  77.46  &  36.73  &  48.68  &  28.66  \\
(5)~~~Removing shuffling ``by balanced pairs'', standard random shuffling instead &    &  57.68 $\pm$0.14   &  77.25  &  36.46  &  48.58  &  28.53  \\
(6)~~~Removing GloVe embeddings to encode the question; learned from random init. instead &    &  56.97 $\pm$0.22   &  75.92  &  36.03  &  48.26  &  27.50  \\
(7a)~~Replace gated tanh layers with tanh activations &    &  52.13 $\pm$7.33   &  75.90  &  34.51  &  38.92  &  24.78  \\
(7b)~~Replace gated tanh layers with ReLU activations &    &  55.38 $\pm$0.05   &  74.65  &  34.83  &  46.33  &  25.24  \\

(8a)~~Binary ground truth targets $s'_{ij}=(s_{ij}\stackrel{?}{>}0.0)$; ~~ReLU &    &  54.34 $\pm$0.19   &  73.81  &  34.50  &  44.95  &  23.75  \\
(8b)~~Binary ground truth targets $s'_{ij}=(s_{ij}\stackrel{?}{=}1.0)$; ~~ReLU &    &  53.41 $\pm$0.20   &  73.23  &  34.38  &  43.54  &  24.32  \\

(9)~~~Replace sigmoid output with softmax; single binary target $s'_{ij}=(s_{ij}\stackrel{?}{=}1.0)$; ~~ReLU &    &  52.52 $\pm$0.31   &  72.79  &  34.33  &  42.09  &  23.81  \\

\Xhline{1\arrayrulewidth}
\end{tabularx}
\normalsize
\vspace{9pt}
\caption{Cumulative ablations of a single network, evaluated on the VQA v2 validation set. The ablations of table rows are cumulative from top to bottom. The experimental setup is identical to the one used for Table~\ref{tableAblative}.}
\label{tableCumul}
\end{table*}

We present an extensive set of experiments that compare our model as presented above (referred to as the \emph{reference} model) with alternative architecture and hyperparameter values. The objective is to show that the proposed model corresponds to a local optimum in the space of architectures and parameters, and to evaluate the sensitivity of the final performance to each design choice. The following discussion follows the structure of Tables~\ref{tableAblative}~and~\ref{tableCumul}. Note the significant breadth and exhaustivity of the following experiments, which represent more than 3,000~GPU-hours of training time.

\paragraph{Experimental setup} Each experiment in this section uses a single network (\ie no ensemble) that is a variation of our \emph{reference} model (first row of Table~\ref{tableAblative}). Each network is trained on the official training set of \emph{VQA v2} and on the additional questions from the \emph{Visual Genome} unless specified. Results are reported on the validation test \emph{VQA v2} at the best epoch (highest overall VQA score). Each experiment (\ie each row of Tables~\ref{tableAblative}~and~\ref{tableCumul}) is repeated 3 times, training the same network with different random seeds. We report the average and standard deviation over those three runs. The main performance metric is the standard VQA accuracy \cite{antol2015vqa}, \ie the average ground truth score of the answers predicted for all questions\footnote{The ground truth score of a candidate answer $j$ to a question $i$ is a value $s_{ij} \in [0,1]$ provided with the dataset. It accounts for possible disagreement between annotators: $s_{ij}=1.0$ if provided by $m\nospace\geq\nospace3$ annotators, and $s=m/3$ otherwise. Those scores are further averaged in a 10--\textit{choose}--9 manner~\cite{antol2015vqa}.}. We additionally report the metric of \emph{Accuracy over pairs}~\cite{zhang2015balanced} (last column of Tables~\ref{tableAblative} and~\ref{tableCumul}). It is the ratio of balanced questions (\ie questions associated with two images leading to two different answers) that are answered perfectly, \ie with both predicted answers having a ground truth score of $1.0$. This metric is significantly harder than the standard per-question score since it requires a correct answer to both images of the pair, discouraging blind guesses and reliance on language priors~\cite{goyal2016balanced,zhang2015balanced}.

We now discuss the results of each ablative experiment from Tables~\ref{tableAblative}~and~\ref{tableCumul} in turn.

\subsection{Training data}
\label{sec:ablativeTrData}

The use of additional training questions/answers from the \emph{\textbf{Visual Genome}} (VG)~\cite{krishnavisualgenome} increase the performance on \emph{VQA v2}~\cite{goyal2016balanced} in all question types. As mentioned above, we only use VG instances with a correct answer appearing the output vocabulary determined on \emph{VQA v2}, and which use an image also used in \emph{VQA v2}. Note that the $+0.67\%$ increase in performance is modest relative to the amount of additional training questions (an additional ~485,000 over the ~650,000 of \emph{VQA v2}). Note that including VG questions relating to MS COCO images not used in \emph{VQA v2} resulted in slightly lower final performance (not reported in Table~\ref{tableAblative}). We did not further investigate this issue.

We compare the proposed \textbf{shuffling of training data} which keeps balanced pairs in the same mini-batches, with a standard, arbitrary random shuffling. The results in Table~\ref{tableAblative} are inconclusive: the overall VQA score is virtually identical either way. The accuracy over pairs however improves with the proposed shuffling. This is to be expected, as the purpose of the proposed method is to improve the learning of differentiating between balanced pairs. The cumulative ablation (row $(5)$ in Table~\ref{tableCumul}) confirms this advantage more clearly.

We evaluate discarding the training questions that do not have their ground truth answer within the selected candidates. Early experiments have shown that those instances do still carry a useful training signal by drawing the predicted scores of the selected candidate answers towards zero. In Table~\ref{tableAblative}, the VQA score remains virtually identical, but a very slight benefit is observed on the accuracy over pairs by retaining those instances.

\subsection{Question embedding}

Our reference model uses pretrained GloVe word embeddings of dimension 300 followed by a one-layer GRU processing words in forward order. We compare this choice to a series of more simple and more advanced options.

Learning the word embeddings from scratch, \ie from random initializations reduces performance by $0.87\%$. The gap with pretrained embeddings is even greater as the model is trained on less training data~(\fig\ref{fig:trData} and \sect\ref{sec:trData}). This is consistent with findings previously reported in~\cite{teney2016zsvqa}. This experiment shows, on the one hand, a benefit of leveraging non-VQA training data. On the other hand, it suggests that a sufficiently large VQA training set may remove this benefit altogether. Using GloVe vectors of smaller dimension ($100$ or $200$) also give lower performance than those in the reference model ($300$).

We investigate whether the pretrained word embeddings (GloVe vectors) really capture word-specific information. The alternative, null hypothesis is that the simple spreading of vectors in the word embedding space is a benefit in itself. To test this, we randomly shuffle the same GloVe vectors, \ie we associate them with words from the input dictionary chosen at random. The shuffled GloVe vectors perform even worse than word embeddings learned from scratch. This shows that GloVe vectors indeed capture word-specific information.

In other experiments (not reported in Table~\ref{tableAblative}), we experimented with a tanh activation following the word embeddings as in \cite{fukui2016multimodal,kazemi2017baseline}. This had no significant effect, either with random or pretrained initializations.

Replacing our GRU with advanced options (backward, bidirectional, or two-layer GRU) gives lower performance. Simpler options (bag-of-words, simply summing or averaging word embeddings) also give lower performance but are still a surprisingly strong baseline, as previously reported in \cite{jabri2016revisiting,teney2016zsvqa}.

\subsection{Image features}

Our best model uses the \textbf{images features from bottom-up attention} of Anderson \etal~\cite{anderson2017features}. These are obtained through a Faster R-CNN framework and an underlying 101-layer ResNet that focuses on specific image regions. The method uses a fixed threshold on object detections, and the number of features $K$ is therefore adaptive to the contents of the image. It is capped at a maximum of 100, and yields an average in the order of $K$=$60$. We experiment with a fixed number of features $K$=$36$. The performance degrades only slightly. This option may be a reasonable alternative considering the lower implementation and computational costs.

We also experiment with mainstream image features from a 200-layer ResNet~\cite{he2015resnet,he2016identityresnet}. As per the common practice, we use the last feature map of the network, of dimensions \nbyn{14} and 2,048 channels. The performance with ResNet features drops dramatically from $63.15\%$ to $57.52\%$. A global average pooling of those features, \ie collapsing to a \nbyn{1} map and discarding the attention mechanism, is expectedly even worse. A surprising finding however is that \textbf{coarser \nbynloose{7} ResNet feature maps} give a reasonable performance of 59.24. Those are obtained by linear interpolation of the \nbyn{14} maps, equivalent to a \nbyn{2} average pooling of stride 2. As an explanation, we hypothesize that the resolution of the coarser maps better correspond to the scale of objects in our images.

Note that the proposed model was initially developed with standard ResNet features, and is not specifically optimized for the features of \cite{anderson2017features}. On the contrary, we found that optimal choices were stable across types of image features (including for the attention mechanism, see Section~\ref{ref:attention}). In all cases, we also observed that an $L2$ normalization of the image features was crucial for good performance~--~at least with our choices of architecture and optimizer.

\subsection{Image attention}
\label{ref:attention}
Our reference model uses a single set of $K$ attention weights normalized with a softmax. Some previous studies have reported  better performance with multiple sets of attention weights (also known as multiple ``glimpses'' or ``attentions heads'') \cite{fukui2016multimodal} and/or with a sigmoid normalization over the weights~\cite{teney2016graphvqa} (which allows multiple glimpses in \emph{one} pass). In our case, none of these options proved beneficial, whether with ResNet features or with features from bottom-up attention. This confirms that the latter image features are suited to a simple drop-in replacement in lieu of traditional feature maps, as argued in~\cite{anderson2017features}.

\subsection{Output vocabulary}

We determine the output vocabulary, \ie the set of candidate answers from those appearing in the training set more than $\ell$ times. We cross-validate this threshold and find a relatively broad optimum around $\ell=8$--$12$ occurrences. This corresponds to about $N$=$2,400$ to $3,800$ candidate answers, which is of the same order as the values typically reported in the literature (without cross-validation) of $2,000$ or $3,000$. Note however a higher $\ell$ (\ie lower $N$) still gives reasonable performance with a lower number of parameters and computational complexity.

\subsection{Output classifier}
\label{sec:classifiers}

Our reference model uses sigmoid outputs and the soft scores $s_ij$ as ground truth targets. We first compare the soft scores with two binarized versions:
\abovedisplayskip=2pt
\belowdisplayskip=6pt
\begin{subequations}
  \begin{align}
  s'_{ij}=(s_{ij}\stackrel{?}{>}0.0) \\
  s'_{ij}=(s_{ij}\stackrel{?}{=}1.0) ~~.
  \end{align}
\end{subequations}
The proposed \textbf{soft scores perform significantly better} than either of the binarized versions~(Table~\ref{tableCumul}).

We then compare the sigmoid outputs of our reference model to the common choice of a softmax. Both use a cross-entropy loss. The softmax uses the single ground truth answer provided in the dataset, whereas the sigmoid uses the complete annotation data, occasionally with multiple correct answers marked for one question, due to disagreement between multiple annotators. The \textbf{sigmoid performs significantly better than a softmax}. This observation is confirmed in the cumulative ablations (Table~\ref{tableCumul})

We now evaluate the proposed pretraining of the parameters $\bw_o^{text}$ and/or $\bw_o^{img}$ of the classifiers. We consider two baselines for those two matrices: random initialization and random shuffling. The former simply initializes and learns the weights like any others in the network. The latter uses the proposed pretraining as initializations but assigns then to the candidate answers randomly by shuffling the rows. This should reveal whether the proposed initialization provides answer-specific information, or whether it simply helps the numerical optimization because of better conditioned initial values. The \textbf{proposed initialization performs consistently better than those baselines}. The random initialization of $\bw_o^{text}$ may seem surprisingly good as it even surpasses the reference model on the overall VQA score. It does not, however, on the metric of \emph{Accuracy over pairs}. The experiments with reduced training data (Section~\ref{sec:trData} and \fig\ref{fig:trData}) confirm even more clearly the benefits of the proposed initialization.

\begin{figure*}[t]
  \centering
  \renewcommand{\tabcolsep}{0.0em}
  \renewcommand{\arraystretch}{0.0}
  \colorbox{white}{
  \begin{tabular}{lllr}\colorbox{white}{\includegraphics[height=0.485\textwidth]{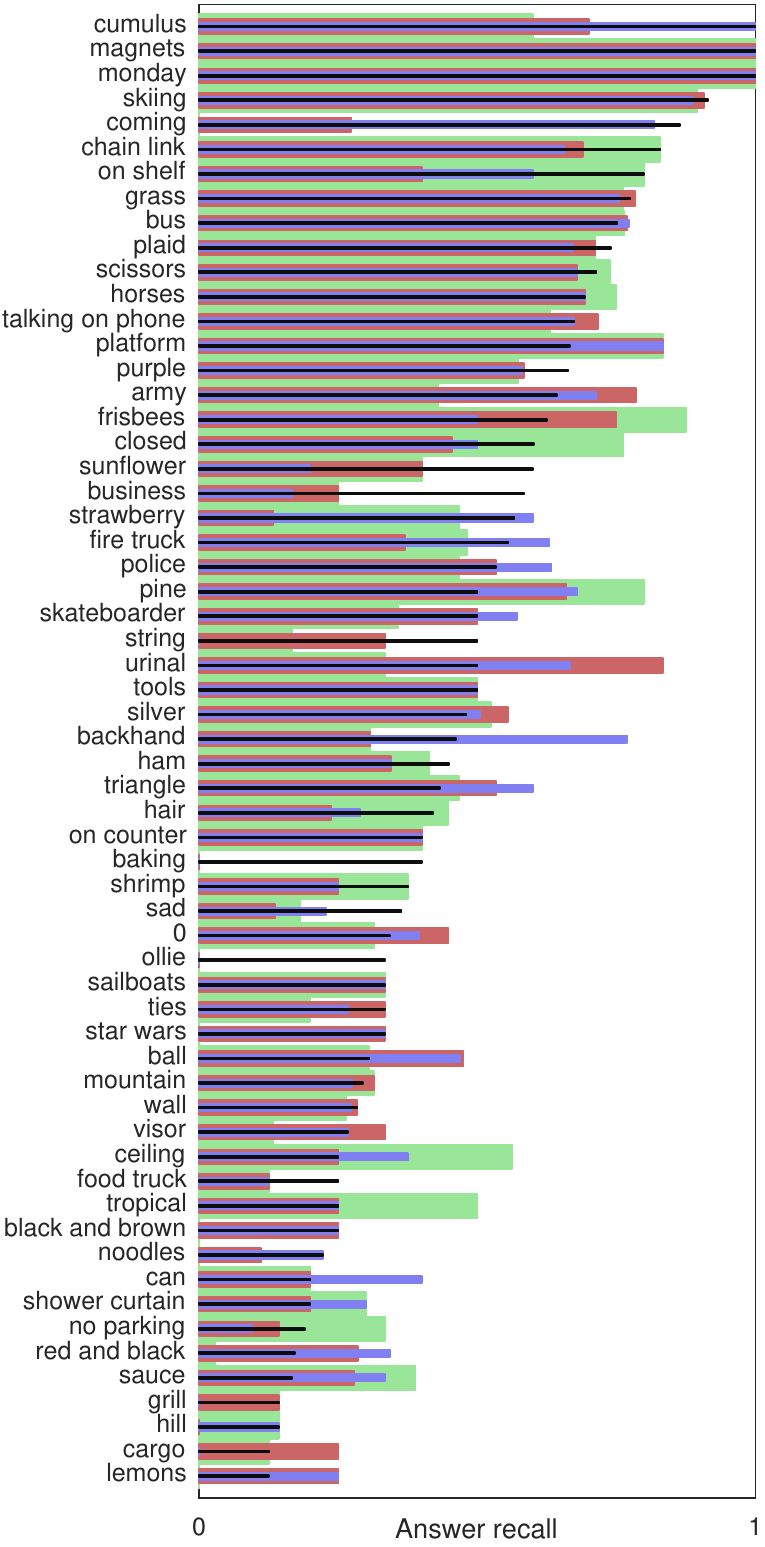}~~}&\colorbox{white}{\includegraphics[height=0.485\textwidth]{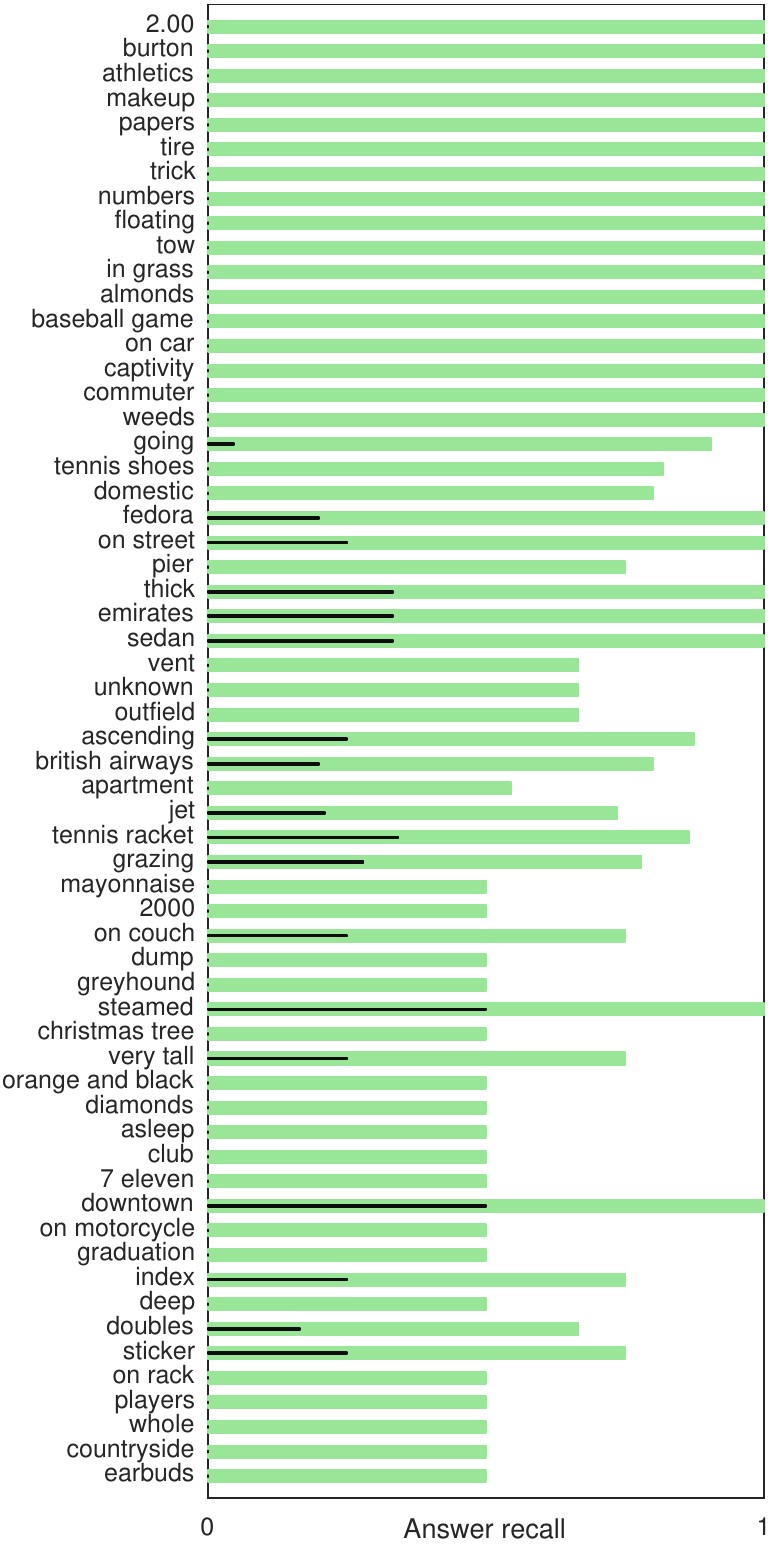}~~}&\colorbox{white}{\includegraphics[height=0.485\textwidth]{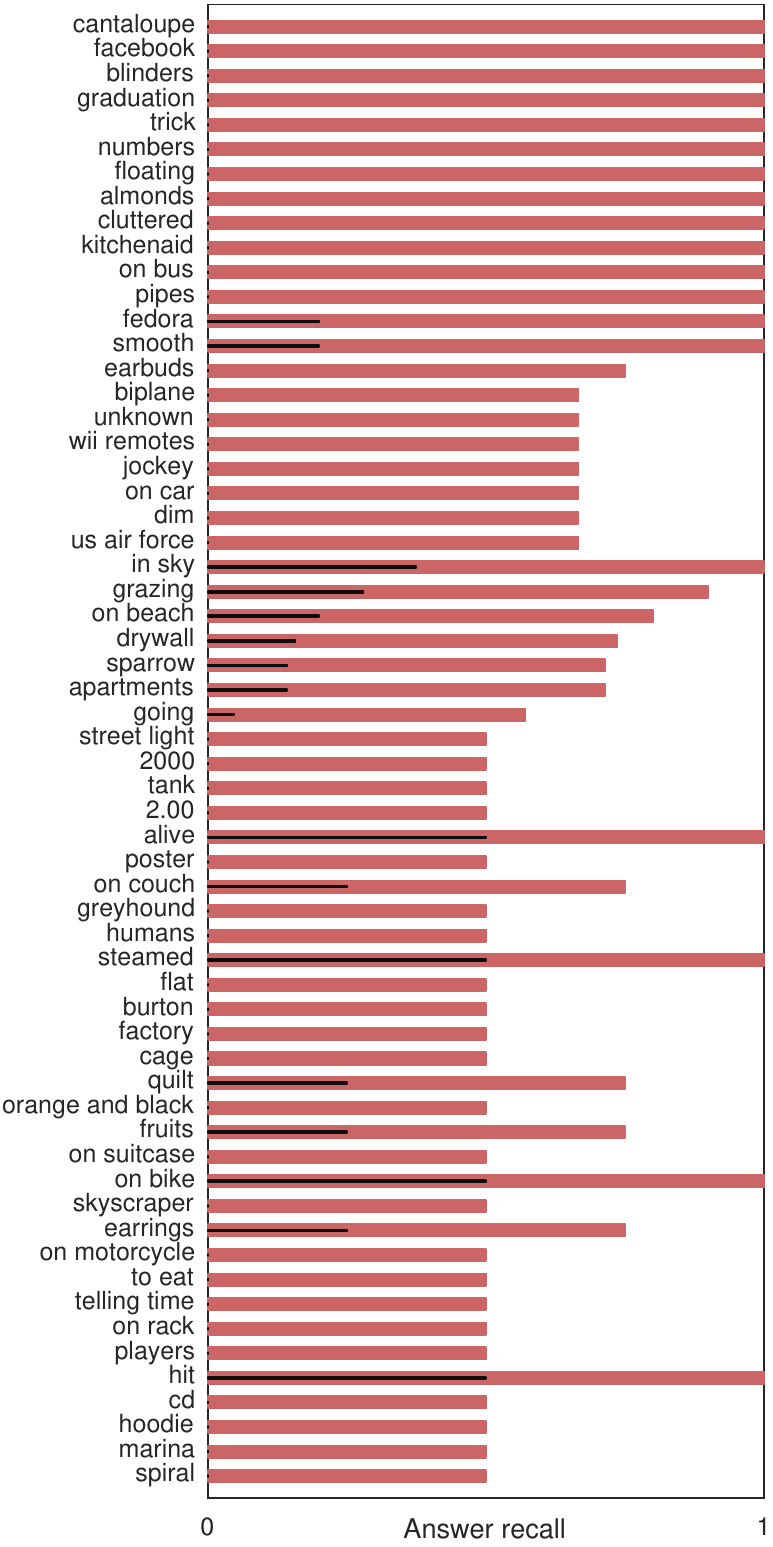}~~}&\colorbox{white}{\includegraphics[height=0.485\textwidth]{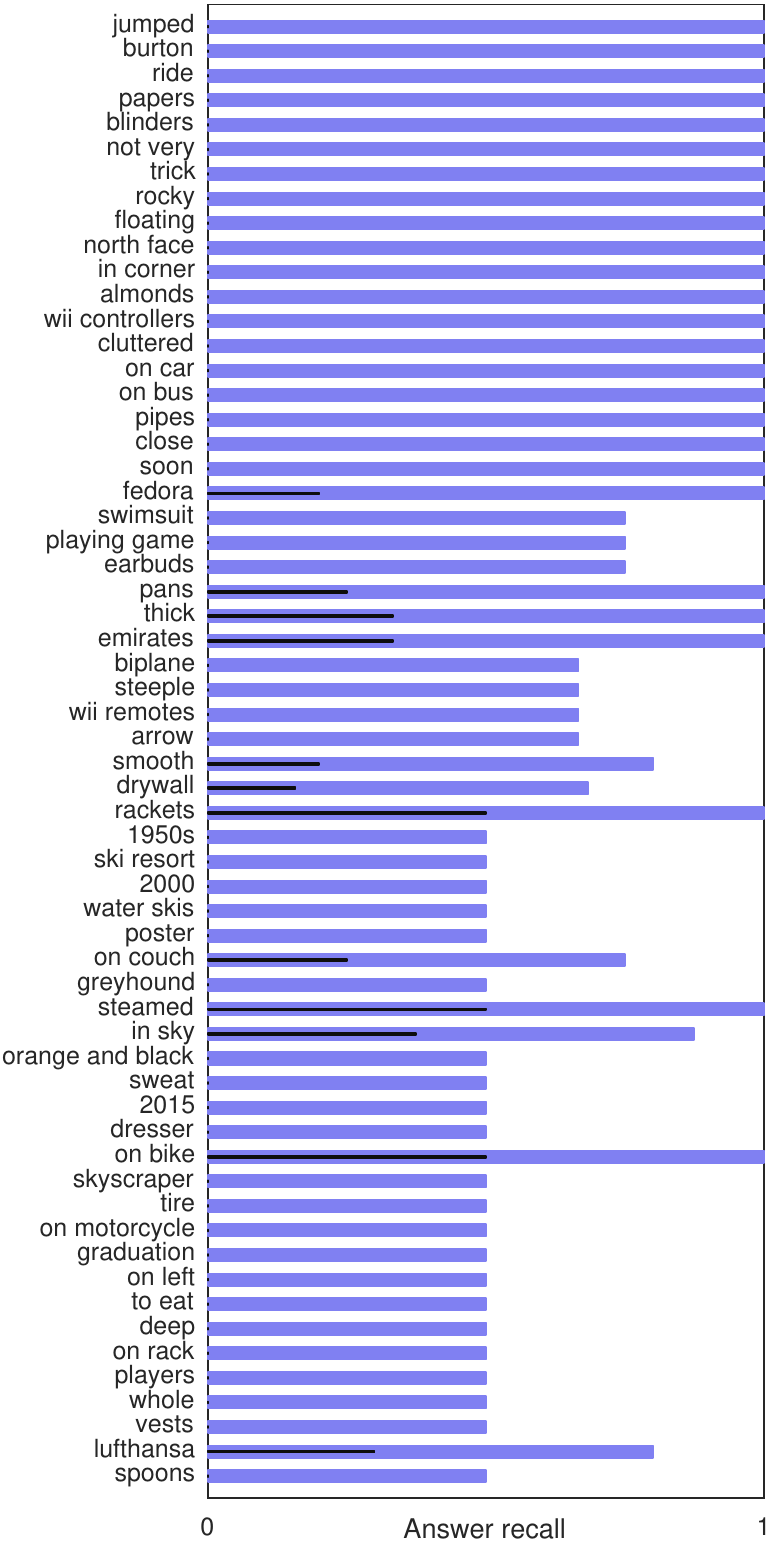}}
  \end{tabular}}\\
  \vspace{8pt}
  \includegraphics[width=0.75\textwidth]{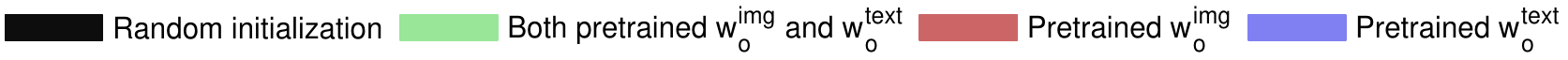}
  \vspace{4pt}
  \label{fig:answers}
  \caption{Effect of pretraining the output classifier on specific answers. We compare the per-candidate-answer recall of three models: a baseline using a classifier trained from scratch or pretrained (in \textbf{black}), and models using pretrained $\bw_o^{text}$ and/or $\bw_o^{img}$. \textit{(Leftmost~chart)}~Random selection of answers sorted by their recall in the baseline model.~~\textit{(Right~three~charts)}~Top-60 answers with the largest improvement in recall by pretraining the classifier. See discussion in Section~\ref{sec:classifiers}.}
\end{figure*}

We take a closer look at the actual answers that improve with the proposed initialization (\fig\ref{fig:answers}). We look at the recall of each candidate answer $j$, defined as
\begin{align}
  \label{eq:recall}
  \mathrm{recall}_i ~=~ \frac{\sum_{i}^{M} (s_{ij}\stackrel{?}{=}1.0 \;~\land~\; \hat{s}_{ij}\stackrel{?}{=}1.0)}{\sum_{i}^{M} (s_{ij}\stackrel{?}{=}1.0)}
\end{align}
where $M$ is the number of evaluated questions, $s_{ij}$ is a ground truth score, and $\hat{s}_{ij}$ a predicted score. In \fig\ref{fig:answers}, we plot the recall of a random selection of answers with and without pretraining the classifier. It is expected that pretraining $\bw_o^{text}$ improves a variety of answers while $\bw_o^{img}$ improves those with clearer visual representation. That intuition is hard to evaluate subjectively and is easy to confirm or disprove by cherry-picking examples. Note that, despite the overall benefit for the proposed approach, the \textbf{recall of many answers is affected negatively}. Other architectures may be necessary to obtain the full benefits of the approach. Another observation --~not directly inferable from \fig\ref{fig:answers}~-- is that the recall the most influenced by pretraining --~positively or negatively~-- is of answers with few training occurrences. This confirms the potential of the approach for \textbf{better handling rare answers}~\cite{teney2016zsvqa}.


\subsection{General architecture}
\label{sec:expArchitecture}

All of our non-linear layers are implemented as \textbf{gated tanh} (Section~\ref{sec:nonLinearLayers}). These show a clear benefit over the gated ReLU, and even more so over simple ReLU or tanh activations. Note that we also experimented, without success, with various other combinations of highway~\cite{srivastava2015highway}, residual~\cite{he2015resnet} and gating connections (not reported in Table~\ref{tableAblative}). One benefit of gated layers is to double the number of learned parameters without increasing the dimension of the hidden states.

We cross-validate the \textbf{dimension of hidden states} among the values~$\{256,384,\underline{512},768,1024,1280\}$. We settled on 512 as a reasonable sweet spot. Larger dimensions (\eg 1280) can be better but without guarantees. The variance across repeated experiments is larger, likely due to overfitting and unstable training.

Our architecture uses a simple \textbf{element-wise product to combine} the question and image representations. This proved far superior to a concatenation (not reported in Table~\ref{tableAblative}), but we did not experiment with the various advanced forms of pooling proposed in the recent literature~\cite{fukui2016multimodal,younes2017mutan}

\subsection{Mini-batch size}

The size of mini-batches during the optimization proves to have a strong influence on the final performance. Mid-range values in~$\{128,\underline{256},\underline{384},\underline{512},768\}$ proved superior to smaller mini-batches (including even smaller values), although they require significantly more memory and high-end GPUs. We observed the optimum mini-batch size to be stable across variations of the network architecture, through other experiments not reported in Table~\ref{tableAblative}.

\subsection{Training set size}
\label{sec:trData}

\begin{figure}[t]
  \centering
  \includegraphics[width=0.90\linewidth]{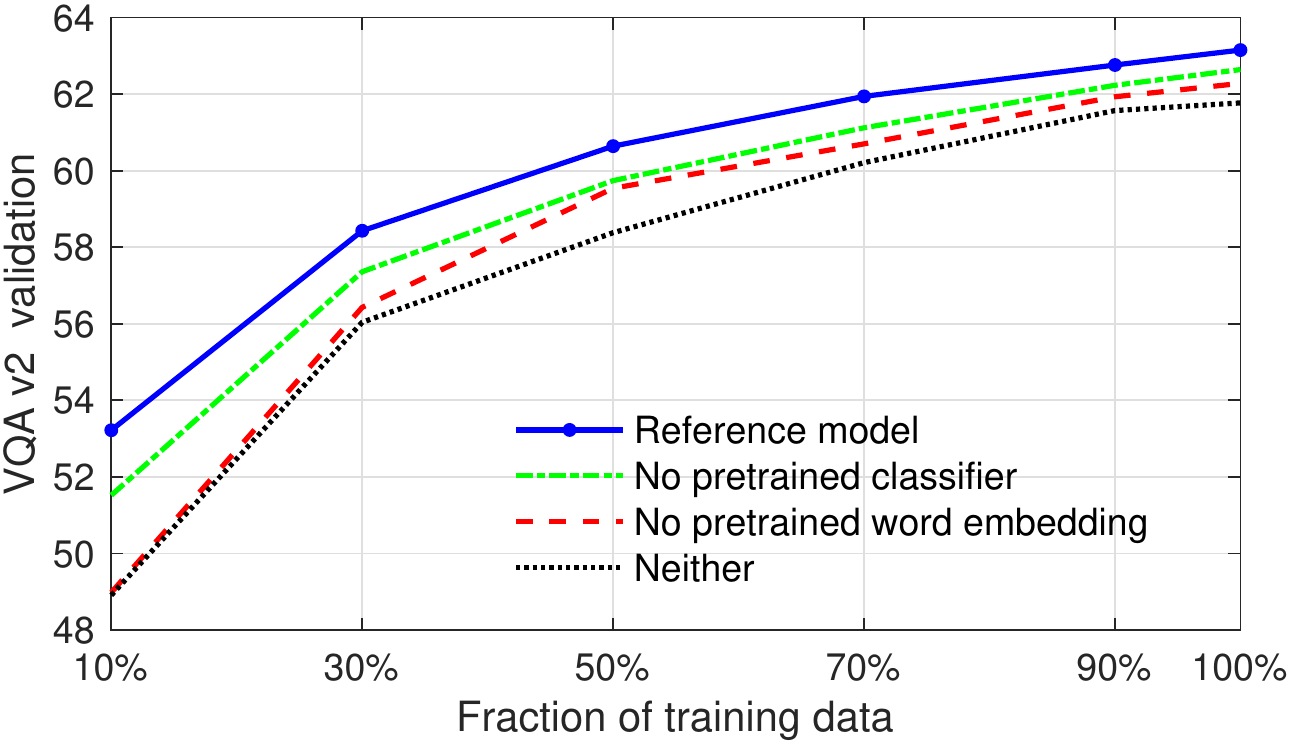}
  \label{fig:trData}
  \caption{Performance of our reference model (Table~\ref{tableAblative}, first row) trained on a subset of the training data. The use of additional non-VQA data for pretraining the word embeddings and the output classifiers is significant, especially when training on a reduced training set. Also not the tendency performance to plateau, and the small gain in performance relative to a 10-fold increase of training data. See discussion in Section~\ref{sec:trData}.}
\end{figure}

We investigate the relation between performance and the quantity of training data. We create random subsets of our training data and train four different models on it.
\vspace{-5pt}
\setlist[enumerate]{leftmargin=*}
\begin{enumerate}[label={\textit{(\arabic*)}}]
  \setlength\itemsep{-4pt}
  \leftmargin=0pt
  \itemindent=0pt
  \listparindent=-10pt
  \topsep=-4pt
  \item Our best reference model.
  \item The ablation that uses word embeddings learned from scratch, instead of GloVe vectors.
  \item The ablation with the output classifier learned from scratch, instead of pretrained $\bw_o^{\textrm{text}}$ and $\bw_o^{\textrm{img}}$.
  \item The conjunction of \textit{(2)} and \textit{(3)}.
\end{enumerate}
\vspace{-4pt}
We plot in \fig\ref{fig:trData} their performance against the amount of training data and make the following observations.
\setlist[itemize]{leftmargin=*}
\begin{itemize}[label={--}]
  \setlength\itemsep{4pt}
  \leftmargin=-10pt
  \itemindent=0pt
  \listparindent=0pt
  \topsep=4pt

  \item Unsurprisingly, the performance improves monotonically with the amount of training data. It roughly follows a logarithmic trend. Remarkably, we already obtain \textbf{reasonable performance with only 10$\%$ of the data}. Consequently, the gain when training on the whole dataset appears small relative to the ten-fold increase in data. That observation is common among natural language tasks in which the data typically follows a Zipf law~\cite{adamic2002zipf} and in other domains with long-tail distributions. In those cases, few training examples are sufficient to learn the most common cases, but an exponentially larger dataset is required for covering more and more of the rare concepts.

  \item The use of \textbf{extra data to pretrain} word embeddings and classifiers is \textbf{always beneficial}. The gap with the baselines models learned from scratch shrinks as more VQA-specific training data is used. It could suggest that a sufficiently large VQA training set would remove the benefit altogether. An alternative view however, is that those other sources of information are most useful for representing rare words and concepts~\cite{teney2016zsvqa} which would require an impractically large dataset to be learned from VQA-specific examples alone. That view then suggests that extra data is necessary in practice.

  \item Pretrained word embeddings and pretrained classifiers each provide a benefit of the same order of magnitude. Importantly, the two techniques are clearly \textbf{complementary} and the best performance is obtained by combining them.
\end{itemize}

\subsection{Ensembling}
\label{sec:ensemble}

\begin{figure}[]
  \centering
  \includegraphics[width=0.90\linewidth]{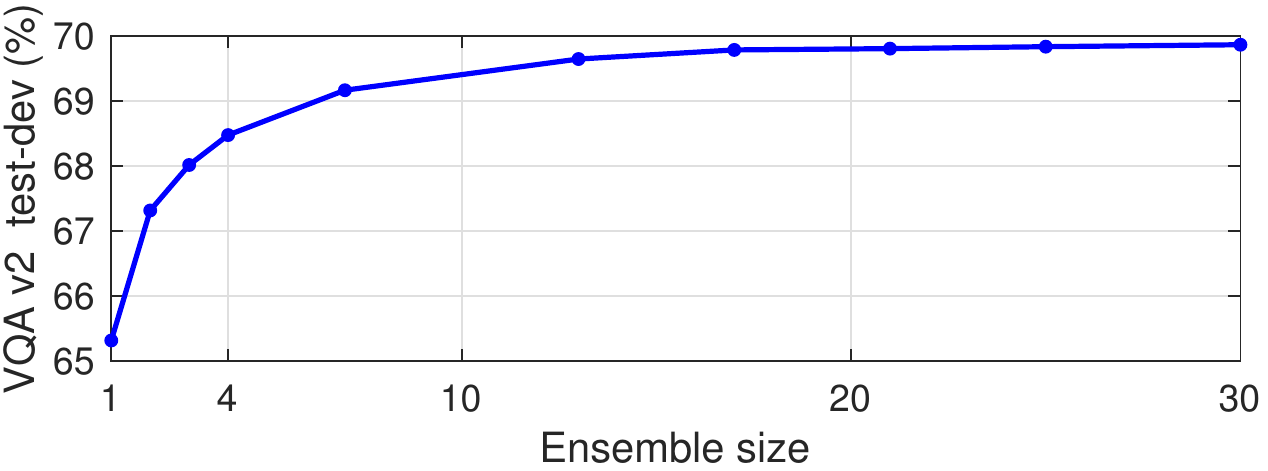}
  \caption{Performance of our best model (last row of Table~\ref{tableLeaderboard}) as a function of the ensemble size. The ensemble uses several instances of a same network trained with different random seeds. Their predicted scores are combined additively. Even small ensembles provide a significant increase in performance over a single network.}
  \label{fig:ensemble}
\end{figure}

We use the common practice of ensembling several networks to obtain better performance. We use the most basic form of ensembling: multiple instances of the same model (same network architecture, same hyperparameters, same data) is trained with different initial random seeds. This affects the initialization of the learned parameters and the stochastic gradient descent optimization. At test time, the scores predicted for the candidates answers by all instances are summed, and the final answer is determined from the highest summed score.

As reported in \fig\ref{fig:ensemble}, the performance increases monotonically with the size of the ensemble, \ie the number of network instances. We obtained our final, best results, with an ensemble of 30 networks. The training of multiple instances is independent and obviously parallelizeable on multiples CPUs or GPUs. Interestingly, \textbf{even small ensembles} of $2$--$5$ instances provide a significant increase in performance over a single network.

Note that those experiments include the validation split of \emph{VQA v2} for training and use its \textit{test-dev} split for evaluation, hence the higher overall performance compared to Tables~\ref{tableAblative}~and~\ref{tableCumul}.


\section{Cumulative ablations}

All ablative experiments presented above consider one or two modifications of the reference model at a time. It is important to note that the cumulative effect of several modifications is not necessarily additive.
In practice, this complicates the search for optimal architectures and hyperparameters. Some choices that appear promising at first may not pan out when combined with other optimizations. Conversely, some options discarded early on the search may prove effective once other hyperparameters have been tuned.

We report in Table~\ref{tableCumul} a series of cumulative ablations of our reference model. We consider a series of characteristics of our model in the inverse of the order in which they could be incorporated into other VQA models. The results follow the trends observed with the individual ablations. Removing each proposed contribution steadily decreases the performance of the model. This set of experiments reveals that the \textbf{most critical components} of our model are the \textbf{sigmoid outputs} instead of a softmax, the \textbf{soft scores} used as ground truth targets, the \textbf{image features from bottom-up attention}~\cite{anderson2017features}, the \textbf{gated tanh activations}, the \textbf{output layers} initialized using GloVE and Google Images, and the \textbf{smart shuffling} of training data.

\section{Comparison with existing methods}

\rowcolors{4}{gray!25}{}

\begin{table*}[t]
\small
\footnotesize
\renewcommand{\arraystretch}{1.34}
\centering`
\begin{tabularx}{\textwidth}{ll c *9{>{\Centering\arraybackslash}X}}
\Xhline{1\arrayrulewidth}
\hiderowcolors
                                          &    &  \multicolumn{4}{c}{VQA v2 test-dev} &  & \multicolumn{4}{c}{VQA v2 test-std} \\ \cline{3-6} \cline{8-11} 
Method                                    &    &  All & Yes/no    &  Numb.   &  Other   &  &  All &  Yes/no  &  Numb.   &  Other \\
\showrowcolors
\Xhline{1\arrayrulewidth}
Prior (most common answer in training set) \cite{goyal2016balanced} &    &  --   &  --   &  --    &  --   &    & 25.98 & 61.20 & ~0.36 & 1.17 \\
LSTM Language only (blind model) \cite{goyal2016balanced} &    &  --   &  --   &  --    &  --   &    & 44.26 & 67.01 & 31.55 & 27.37 \\
Deeper LSTM Q norm. I \cite{lu2015deeperLstm} as reported in \cite{goyal2016balanced} &    &  --   &  --   &  --    &  --   &    & 54.22 & 73.46 & 35.18 & 41.83 \\
MCB \cite{fukui2016multimodal} as reported in \cite{goyal2016balanced} &    &  --   &  --   &  --    &  --   &    & 62.27 & 78.82 & 38.28 & 53.36 \\
UPMC-LIP6 \cite{younes2017mutan} &    &  --   &  --   &  --    &  --   &    & 65.71 & 82.07 & 41.06 & 57.12 \\
Athena &    &  --   &  --   &  --    &  --   &    & 67.59 & 82.50 & 44.19 & 59.97 \\
LV-NUS &    &  --   &  --   &  --    &  --   &    &  66.77 & 81.89 & 46.29 & 58.30  \\
HDU-USYD-UNCC &    &  --   &  --   &  --    &  --   &    & 68.09 & 84.50 & 45.39 & 59.01 \\
\hline
\hiderowcolors
\multicolumn{11}{l}{Proposed model} \\
\showrowcolors
\hspace{1.0em} ResNet features \nbyn{7}, single network  & & 62.07 & 79.20 & 39.46 & 52.62 & & 62.27 & 79.32 & 39.77 & 52.59 \\
\hspace{1.0em} Image features from bottom-up attention, adaptive $K$, single network & & 65.32 & 81.82 & 44.21 & 56.05 & & 65.67 & 82.20 & 43.90 & 56.26 \\
\hspace{1.0em} ResNet features \nbyn{7}, ensemble  & & 66.34 & 83.38 & 43.17 & 57.10  &  & 66.73 & 83.71 & 43.77 & 57.20 \\
\hspace{1.0em} \textbf{Image features from bottom-up attention, adaptive $K$, ensemble}  & & \textbf{69.87} & \textbf{86.08} & \textbf{48.99} & \textbf{60.80} & & \textbf{70.34} & \textbf{86.60} & \textbf{48.64} & \textbf{61.15} \\
\Xhline{1\arrayrulewidth}
\end{tabularx}
  \normalsize
  \vspace{9pt}
  \caption{Comparison of our best model with competing methods. Excerpt from the official \emph{VQA v2 Leaderboard}~\cite{vqaleaderboard}.}
  \label{tableLeaderboard}
\end{table*}


We compare in Table~\ref{tableLeaderboard} the performance of our best model with existing methods. Ours is an ensemble of 30 networks identical to the \emph{reference} model (first row of Table~\ref{tableAblative}) with the exception of the dimension of the hidden states, increased here to $1,500$. The issue of overfitting (Section~\ref{sec:expArchitecture}) is mitigated by the large ensemble size. Compared to Table~\ref{tableAblative}, this model also includes here the validation split of \emph{VQA v2} for training. Our model obtained the first place at the 2017 VQA Challenge~\cite{vqaleaderboard}. It still surpasses all competing methods by a significant margin at the time of writing.


\section{Discussion and conclusions}

This paper presented a model for VQA based on a deep neural network that significantly outperforms all other approaches proposed to date. Importantly, we reported an extensive suite of experiments that identify the contribution of each design choice and the performance of alternative designs. The general take-away from this study is that the performance is very dependent on design choices and on various details of the implementation. We attribute the success of our model to a number of points. Some are seemingly minor and easily implemented (\eg large mini-batch size, sigmoid output) while others are clearly non-trivial (\eg gated non-linear activations, image features from bottom-up attention, pretraining the output classifier).

This paper does not claim to make breakthrough advances in the field of VQA, which remains a largely unsolved problem. We hope that our model may however serve as a solid basis on which to make future progress. Our extensive analysis and exploration of designs is unprecedented in scale for VQA, and is intended as a significant contribution to the field. It provides indicators on the importance of various components of a VQA model. It also allows us to point at several promising directions for future developments.

We showed that significant gains were still achievable through better image features, in particular with the region-specific features of \cite{anderson2017features} that use bottom-up attention. We also measured that gains from additional VQA training data had not reached a clear plateau yet. However, the trend of collecting larger datasets is unlikely to bring significant breakthroughs. We believe that incorporating other sources of information and leveraging non-VQA datasets is a promising direction.

Our evaluation of simple baselines have shown surprisingly strong performances. For example, encoding questions as a simple bag-of-words performs almost as well as state-of-the-art recurrent encoders. It suggests that the word ordering in questions may not convey much information~--~which is plausible for the most basic ones~--~or, more realistically, that our current models are still unable to understand and make effective use of language structure. Recent works on compositional models for VQA are a promising direction to address this issue~\cite{andreas2016learning,hu2017reason,johnson2016clevr,johnson2017programs}.

Finally, we must be reminded to look at performance measures with a critical eye. The commonly reported metrics such as our $>\nospace70\%$ per-question accuracy appear encouraging, but it is valuable to keep an eye on failure cases and alternative performance measures. We reported throughout this study the \emph{accuracy over balanced pairs} of questions. This stricter measure requires accurate answers to two complimentary versions of a same question relating to different images. It better reflects the ability of the method for visual understanding and for making out subtle differences between images. In that case, the performance drops to the order $35\%$. While far less impressive, that figure is more representative of our current state of progress on VQA. We hope that keeping such a critical outlook will encourage more radical innovation and breakthroughs in the near future.


\newpage

{\small\bibliographystyle{ieee}\bibliography{Bibliography}}

\begin{thebibliography}{10}\itemsep=-1pt

\bibitem{vqaleaderboard}
{VQA Challenge leaderboard}.
\newblock \url{http://visualqa.org} and \url{http://evalai.cloudcv.org}.

\bibitem{adamic2002zipf}
L.~A. Adamic and B.~A. Huberman.
\newblock Zipfs law and the internet.
\newblock {\em Glottometrics}, 3(1):143--150, 2002.

\bibitem{anderson2017features}
P.~Anderson, X.~He, C.~Buehler, D.~Teney, M.~Johnson, S.~Gould, and L.~Zhang.
\newblock Bottom-up and top-down attention for image captioning and vqa.
\newblock {\em arXiv preprint arXiv:1707.07998}, 2017.

\bibitem{andreas2016learning}
J.~Andreas, M.~Rohrbach, T.~Darrell, and D.~Klein.
\newblock Learning to compose neural networks for question answering.
\newblock In {\em Annual Conference of the North American Chapter of the
  Association for Computational Linguistics}, 2016.

\bibitem{andreas2015deep}
J.~Andreas, M.~Rohrbach, T.~Darrell, and D.~Klein.
\newblock {Neural Module Networks}.
\newblock In {\em {Proc. IEEE Conf. Comp. Vis. Patt. Recogn.}}, 2016.

\bibitem{antol2015vqa}
S.~Antol, A.~Agrawal, J.~Lu, M.~Mitchell, D.~Batra, C.~L. Zitnick, and
  D.~Parikh.
\newblock {{VQA}: Visual Question Answering}.
\newblock In {\em {Proc. IEEE Int. Conf. Comp. Vis.}}, 2015.

\bibitem{younes2017mutan}
H.~Ben{-}younes, R.~Cad{\`{e}}ne, M.~Cord, and N.~Thome.
\newblock {MUTAN:} multimodal tucker fusion for visual question answering.
\newblock {\em arXiv preprint arXiv:1705.06676}, 2017.

\bibitem{Chen2015ABC}
K.~Chen, J.~Wang, L.-C. Chen, H.~Gao, W.~Xu, and R.~Nevatia.
\newblock {ABC-CNN: An Attention Based Convolutional Neural Network for Visual
  Question Answering}.
\newblock {\em arXiv preprint arXiv:1511.05960}, 2015.

\bibitem{cho2014learning}
K.~Cho, B.~van Merrienboer, C.~Gulcehre, F.~Bougares, H.~Schwenk, and
  Y.~Bengio.
\newblock {Learning phrase representations using {RNN} encoder-decoder for
  statistical machine translation}.
\newblock In {\em {Proc. Conf. Empirical Methods in Natural Language
  Processing}}, 2014.

\bibitem{das2017dialog}
A.~Das, S.~Kottur, K.~Gupta, A.~Singh, D.~Yadav, J.~M. Moura, D.~Parikh, and
  D.~Batra.
\newblock {V}isual {D}ialog.
\newblock In {\em Proceedings of the IEEE Conference on Computer Vision and
  Pattern Recognition}, 2017.

\bibitem{dauphin2016languagecnn}
Y.~N. Dauphin, A.~Fan, M.~Auli, and D.~Grangier.
\newblock Language modeling with gated convolutional networks.
\newblock {\em arXiv preprint arXiv:1612.08083}, 2016.

\bibitem{fang2014captions}
H.~Fang, S.~Gupta, F.~Iandola, R.~Srivastava, L.~Deng, P.~Doll{\'a}r, J.~Gao,
  X.~He, M.~Mitchell, J.~Platt, et~al.
\newblock {From captions to visual concepts and back}.
\newblock In {\em {Proc. IEEE Conf. Comp. Vis. Patt. Recogn.}}, 2015.

\bibitem{fukui2016multimodal}
A.~Fukui, D.~H. Park, D.~Yang, A.~Rohrbach, T.~Darrell, and M.~Rohrbach.
\newblock Multimodal compact bilinear pooling for visual question answering and
  visual grounding.
\newblock {\em arXiv preprint arXiv:1606.01847}, 2016.

\bibitem{goyal2016balanced}
Y.~Goyal, T.~Khot, D.~Summers-Stay, D.~Batra, and D.~Parikh.
\newblock Making the {V} in {VQA} matter: Elevating the role of image
  understanding in {V}isual {Q}uestion {A}nswering.
\newblock {\em arXiv preprint arXiv:1612.00837}, 2016.

\bibitem{he2015resnet}
K.~He, X.~Zhang, S.~Ren, and J.~Sun.
\newblock Deep residual learning for image recognition.
\newblock In {\em {Proc. IEEE Conf. Comp. Vis. Patt. Recogn.}}, 2016.

\bibitem{he2016identityresnet}
K.~He, X.~Zhang, S.~Ren, and J.~Sun.
\newblock Identity mappings in deep residual networks.
\newblock {\em arXiv preprint arXiv:1603.05027}, 2016.

\bibitem{hu2017reason}
R.~Hu, J.~Andreas, M.~Rohrbach, T.~Darrell, and K.~Saenko.
\newblock Learning to reason: End-to-end module networks for visual question
  answering.
\newblock {\em arXiv preprint arXiv:1704.05526}, 2017.

\bibitem{jabri2016revisiting}
A.~Jabri, A.~Joulin, and L.~van~der Maaten.
\newblock Revisiting visual question answering baselines.
\newblock 2016.

\bibitem{Jiang2015compositional}
A.~Jiang, F.~Wang, F.~Porikli, and Y.~Li.
\newblock {Compositional Memory for Visual Question Answering}.
\newblock {\em arXiv preprint arXiv:1511.05676}, 2015.

\bibitem{johnson2016clevr}
J.~Johnson, B.~Hariharan, L.~van~der Maaten, L.~Fei{-}Fei, C.~L. Zitnick, and
  R.~B. Girshick.
\newblock {CLEVR:} {A} diagnostic dataset for compositional language and
  elementary visual reasoning.
\newblock {\em arXiv preprint arXiv:1612.06890}, 2016.

\bibitem{johnson2017programs}
J.~Johnson, B.~Hariharan, L.~van~der Maaten, J.~Hoffman, F.~Li, C.~L. Zitnick,
  and R.~B. Girshick.
\newblock Inferring and executing programs for visual reasoning.
\newblock {\em arXiv preprint arXiv:1705.03633}, 2017.

\bibitem{kazemi2017baseline}
V.~Kazemi and A.~Elqursh.
\newblock Show, ask, attend, and answer: A strong baseline for visual question
  answering.
\newblock {\em arXiv preprint arXiv:1704.03162}, 2017.

\bibitem{Kim2016TrimZero}
J.-H. Kim, J.~Kim, J.-W. Ha, and B.-T. Zhang.
\newblock {TrimZero: A Torch Recurrent Module for Efficient Natural Language
  Processing}.
\newblock In {\em Proceedings of KIIS Spring Conference}, volume~26, pages
  165--166, 2016.

\bibitem{krishnavisualgenome}
R.~Krishna, Y.~Zhu, O.~Groth, J.~Johnson, K.~Hata, J.~Kravitz, S.~Chen,
  Y.~Kalantidis, L.-J. Li, D.~A. Shamma, M.~Bernstein, and L.~Fei-Fei.
\newblock Visual genome: Connecting language and vision using crowdsourced
  dense image annotations.
\newblock {\em arXiv preprint arXiv:1602.07332}, 2016.

\bibitem{lin2014microsoft}
T.-Y. Lin, M.~Maire, S.~Belongie, J.~Hays, P.~Perona, D.~Ramanan,
  P.~Doll{\'a}r, and C.~L. Zitnick.
\newblock {Microsoft COCO: Common objects in context}.
\newblock In {\em {Proc. Eur. Conf. Comp. Vis.}}, 2014.

\bibitem{lu2016hierarchical}
J.~Lu, J.~Yang, D.~Batra, and D.~Parikh.
\newblock Hierarchical question-image co-attention for visual question
  answering.
\newblock {\em arXiv preprint arXiv:1606.00061}, 2016.

\bibitem{pennington2014glove}
J.~Pennington, R.~Socher, and C.~Manning.
\newblock {Glove: Global Vectors for Word Representation}.
\newblock In {\em Conference on Empirical Methods in Natural Language
  Processing}, 2014.

\bibitem{ren2015faster}
S.~Ren, K.~He, R.~Girshick, and J.~Sun.
\newblock Faster r-cnn: Towards real-time object detection with region proposal
  networks.
\newblock In {\em {Proc. Advances in Neural Inf. Process. Syst.}} 2015.

\bibitem{srivastava2015highway}
R.~K. Srivastava, K.~Greff, and J.~Schmidhuber.
\newblock Highway networks.
\newblock {\em arXiv preprint arXiv:1505.00387v1}, 2015.

\bibitem{teney2016graphvqa}
D.~Teney, L.~Liu, and A.~van~den Hengel.
\newblock Graph-structured representations for visual question answering.
\newblock In {\em {Proc. IEEE Conf. Comp. Vis. Patt. Recogn.}}, 2016.

\bibitem{teney2016zsvqa}
D.~Teney and A.~van~den Hengel.
\newblock Zero-shot visual question answering.
\newblock 2016.

\bibitem{vinyals2014show}
O.~Vinyals, A.~Toshev, S.~Bengio, and D.~Erhan.
\newblock {Show and tell: A neural image caption generator}.
\newblock In {\em {Proc. IEEE Conf. Comp. Vis. Patt. Recogn.}}, 2014.

\bibitem{wu2017survey}
Q.~Wu, D.~Teney, P.~Wang, C.~Shen, A.~Dick, and A.~van~den Hengel.
\newblock Visual question answering: A survey of methods and datasets.
\newblock {\em Computer Vision and Image Understanding}, 2017.

\bibitem{xu2015ask}
H.~Xu and K.~Saenko.
\newblock {Ask, Attend and Answer: Exploring Question-Guided Spatial Attention
  for Visual Question Answering}.
\newblock {\em arXiv preprint arXiv:1511.05234}, 2015.

\bibitem{yang2015stacked}
Z.~Yang, X.~He, J.~Gao, L.~Deng, and A.~Smola.
\newblock {Stacked Attention Networks for Image Question Answering}.
\newblock In {\em {Proc. IEEE Conf. Comp. Vis. Patt. Recogn.}}, 2016.

\bibitem{zeiler2012adadelta}
M.~D. Zeiler.
\newblock {ADADELTA:} an adaptive learning rate method.
\newblock {\em arXiv preprint arXiv:1212.5701}, 2012.

\bibitem{zhang2015balanced}
P.~Zhang, Y.~Goyal, D.~Summers{-}Stay, D.~Batra, and D.~Parikh.
\newblock Yin and yang: Balancing and answering binary visual questions.
\newblock In {\em {Proc. IEEE Conf. Comp. Vis. Patt. Recogn.}}, 2016.

\bibitem{zhu2015visual7w}
Y.~Zhu, O.~Groth, M.~Bernstein, and L.~Fei-Fei.
\newblock {Visual7W: Grounded Question Answering in Images}.
\newblock In {\em {Proc. IEEE Conf. Comp. Vis. Patt. Recogn.}}, 2016.

\end{thebibliography}
\clearpage

\end{document}